\documentclass[square,numbers]{article}

\usepackage[final]{neurips_2022}

\usepackage[utf8]{inputenc} 
\usepackage[T1]{fontenc}    
\usepackage{hyperref}       
\usepackage{url}            
\usepackage{booktabs}       
\usepackage{amsfonts}       
\usepackage{nicefrac}       
\usepackage{microtype}      
\usepackage{xcolor}         
\usepackage{graphicx}
\usepackage{makecell}
\usepackage{dsfont}
\usepackage{booktabs}
\usepackage{tikz}
\usetikzlibrary{bayesnet}
\usetikzlibrary{arrows}
\usepackage{color}
\usepackage{caption}
\usepackage{subcaption}
\usetikzlibrary{backgrounds}
\usepackage[ruled]{algorithm2e}
\usepackage{natbib}
\usepackage{mathtools}
\usepackage{esint}

\usepackage{mathtools}
\usepackage{amsfonts}       
\usepackage{amsmath}    
\usepackage{amssymb}    
\usepackage{amsthm}     
\usepackage[export]{adjustbox}
\usepackage{wrapfig}
\newtheorem{theorem}{Result}[section]

\newtheorem{realtheorem}{Theorem}[section]

\newcommand{\method}{CFQP}
\SetKwComment{Comment}{/* }{ */}
\SetKw{KwBy}{by}

\title{Deep Counterfactual Estimation with Categorical Background Variables}

\author{%
  Edward De Brouwer\\
  ESAT-STADIUS\\
  KU Leuven\\
  \texttt{edward.debrouwer@esat.kuleuven.be} \\
}

\begin{document}

\newtheorem{prop}{Proposition}
\theoremstyle{definition}
\newtheorem{definition}{Definition}
\newtheorem{realdefinition}{Definition}[section]
\newtheorem{corollary}{Corollary}[theorem]
\newtheorem{assumption}{Assumption}

\maketitle

\begin{abstract}
  Referred to as the third rung of the causal inference ladder, counterfactual queries typically ask the "\emph{What if ?}" question retrospectively. The standard approach to estimate counterfactuals resides in using a structural equation model that accurately reflects the underlying data generating process. However, such models are seldom available in practice and one usually wishes to infer them from observational data alone. Unfortunately, the correct structural equation model is in general not identifiable from the observed factual distribution. Nevertheless, in this work, we show that under the assumption that the main latent contributors to the treatment responses are categorical, the counterfactuals can be still reliably predicted.
  Building upon this assumption, we introduce CounterFactual Query Prediction (\method), a novel method to infer counterfactuals from continuous observations when the background variables are categorical. We show that our method significantly outperforms previously available deep-learning-based counterfactual methods, both theoretically and empirically on time series and image data. Our code is available at \url{https://github.com/edebrouwer/cfqp}.
\end{abstract}

\section{Introduction}

Counterfactual queries aim at inferring the impact of a treatment conditioned on another observed treatment outcome. 
Typically, given an individual, a treatment assignment, and a treatment outcome, the counterfactual question asks what would have happened to that individual, had it been given another treatment, everything else being equal. An illustrative and motivating example is the case of clinical time series. Based on the observation of the outcome of treatment $A$ on a particular patient, counterfactual queries ask what would have been the outcome for this patient, had it been given treatment $B$ instead. Notably, counterfactual prediction differs from interventional prediction, which is also referred to as counterfactual potential outcomes~\citep{rubin1974estimating} and constitutes the second rung of the causation ladder~\cite{pearl2009causal}. Counterfactual predictions are retrospective, as they condition on an observed treatment outcome. In contrast, interventional predictions are prospective as they only condition on observations obtained before treatment assignment.


Much more than a statistical curiosity, counterfactual reasoning reflects complex cognitive abilities that are deeply ingrained in the human brain \cite{roese2014might} and emerges in the early stages of cognitive development \cite{byrne2016counterfactual}. The ability to reason counterfactually can indeed help to identify causes of outcomes retrospectively, has been suggested to be central in the formation of rational intention, and supports key theories of human cognition \cite{sep-counterfactuals}. The importance of counterfactual reasoning in the human cognitive process has thus motivated researchers to endow artificially intelligent systems with the same ability \cite{pearl2000models}. However, counterfactual inference is not possible from observational and interventional data alone \cite{pearl20217}.

Counterfactual reasoning, therefore, requires making several assumptions to overcome this limitation. The most popular one assumes knowledge of the underlying structural equation model that describes the data generating process~\cite{pearl2000models} or a specific functional form thereof \cite{abadie2021using,balke2013counterfactuals,pearl2000models}. Unfortunately, this assumption is rarely met in practice, especially in high-dimensional data such as time series or images. This led to the development of \emph{deep} structural equation models that attempt to model the structural equations with neural networks \cite{pawlowski2020deep,sanchez2021diffusion}. However, despite their ability to model high-dimensional data, these approaches fail to provide theoretical guarantees for the reconstruction of counterfactuals. Indeed, they focus on modeling the factual distribution which, without further assumption, can, unfortunately, lead to erroneous counterfactual distributions.

In this work, we bridge the gap between the classical structural equation model assumptions and deep-learning-based architectures. By assuming that the treatment and observables are continuous and that the hidden variables that contribute most to the treatment response are categorical, we can rely on recent results in identifiability of mixture distributions \cite{aragam2020identifiability} to show that we can approximately recover the counterfactuals using arbitrary parametric functions (\emph{i.e.} deep neural networks) to model the causal dependence between variables. This allows us to infer counterfactuals on high-dimensional data such as time series and images. 
Generally, this work explores the assumptions that can lead to approximate counterfactual reconstructions while controlling the discrepancy between the recovered and the true counterfactual distributions.

Besides the general appeal of endowing machine learning architectures with counterfactual reasoning abilities, an important motivation for our work is the counterfactual estimation of treatment effects in clinical patient trajectories. In this motivational example, one wishes to predict the individual treatment effect retrospectively. Based on the observed treatment outcome of a particular patient, we want to predict what would have been the outcome under a different treatment assignment. The ability to perform counterfactual inference on patient trajectories has indeed been identified as a potential tool for improving long and costly randomized clinical trials~\cite{nguyen2020counterfactual}.

\vspace{-5pt}

\paragraph*{Contributions}
\begin{itemize}
\itemsep-0.1em
\item We provide a new set of assumptions under which the counterfactuals are identifiable using arbitrary neural networks architecture, bridging the gap between structural equation models and deep learning architectures.
\item We derive a new counterfactual identifiability result that motivates a novel counterfactual reconstruction architecture.
\item We evaluate our construction on three different datasets with different high-dimensional modalities (images and time series) and demonstrate accurate counterfactual estimation.
\end{itemize}
\section{Background}

\subsection{Problem Setup : Counterfactual Estimation}
\label{sec:problem_setup}

\begin{figure}[!htbp]
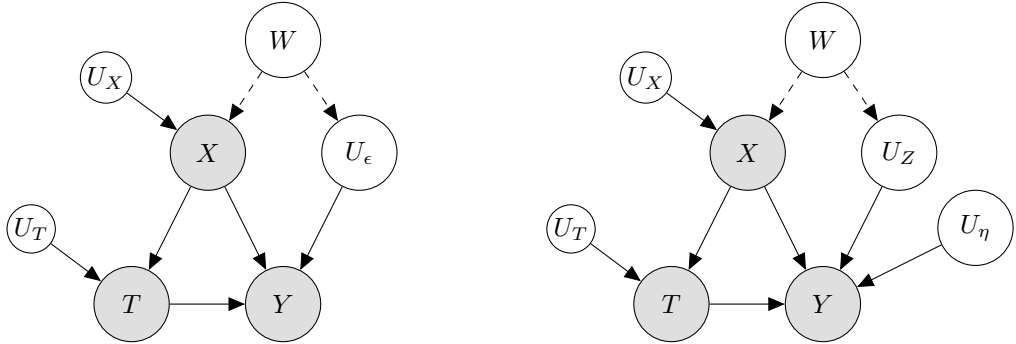

   \begin{subfigure}[t]{0.45\textwidth}
     \centering
  \tikz{
\node[obs,minimum size=1cm] (y) {$Y$};%
\node[obs,above=of y,xshift=-1cm,minimum size=1cm] (x) {$X$};%
\node[obs,left=of y,minimum size=1cm] (t) {$T$};%
\node[latent, right=of x,minimum size=1cm] (e) {$U_{\epsilon}$};%
\node[latent, above=of y,minimum size=1cm,yshift = 1.5cm] (w) {$W$};%
\node[latent,minimum size=0.5cm,left=of x, yshift = 1cm,xshift = 0.5cm] (ux) {$U_X$};%
\node[latent,minimum size=0.5cm,left=of t, yshift = 1cm,xshift = 0.5cm] (ut) {$U_T$};%

 \edge {x,t,e} {y}
 \edge {ux}{x}
  \edge {ut}{t}
 \edge[dashed] {w} {x}
  \edge[dashed] {w} {e}
  \edge {x} {t}}
    \caption{General Bayesian network for the treatment counterfactual problem. $X$, $Y$ and $T$ are observed while $U_X,U_T,W$ and $U_{\epsilon}$ are hidden background variables.}
    \label{fig:counterfactual_dag1}
   \end{subfigure}\hfill
   \begin{subfigure}[t]{0.45\textwidth}
 \centering
  \tikz{
\node[obs,minimum size=1cm] (y) {$Y$};%
\node[obs,above=of y,xshift=-1cm,minimum size=1cm] (x) {$X$};%
\node[obs,left=of y,minimum size=1cm] (t) {$T$};%
\node[latent, right=of x,minimum size=1cm] (z) {$U_Z$};%
\node[latent, right=of z,minimum size=1cm,xshift = -1cm, yshift = -1cm] (n) {$U_{\eta}$};%
\node[latent, above=of y,minimum size=1cm,yshift = 1.5cm] (w) {$W$};%
\node[latent,minimum size=0.5cm,left=of x, yshift = 1cm,xshift = 0.5cm] (ux) {$U_X$};%
\node[latent,minimum size=0.5cm,left=of t, yshift = 1cm,xshift = 0.5cm] (ut) {$U_T$};%
 
\edge {ux}{x}
\edge {ut}{t}
\edge[dashed]{w}{x}
\edge[dashed]{w}{z}
 \edge {x,t,z,n} {y}
  \edge {x} {t}}
    \caption{Bayesian Network embodying the hidden categorical background variable assumption. $U_{\epsilon}$ is split in a background categorical variable $U_Z$ and a continuous background variable $U_{\eta}$.}
    \label{fig:counterfactual_dag2}
   \end{subfigure}
   \caption{Graphical model representations of the causal model $M$. We assume strong ignorability and continuous treatments $T$, observables $X$ and responses $Y$.}
   \label{fig:counterfactual_dags}
\end{figure}

We consider the general causal model $M = \langle U,V,F \rangle$ depicted in Figure \ref{fig:counterfactual_dag1} consisting of background variables $U$, endogenous variables $X,T$ and $Y$ and the set of structural functions $F$. Background variables $U= \{U_X,U_T,U_{\epsilon},W\}$ are hidden exogenous random variables that determine the values of the observed variables $V=\{X,T,Y\}$. Covariates $X \in \mathcal{X}$ represent the information available before treatment assignment, $T \in \mathcal{T}$ is the treatment assignment and $Y \in \mathcal{Y}$ is the observed response to the treatment. We refer to the space of probability measures on $\mathcal{Y}$ as $\mathcal{P}(\mathcal{Y})$. Observed variables $V$ are generated following the structural equations $F = \{f_X,f_T,f_Y,f_{\epsilon}\}$, such that $X = f_X(U_x,W)$, $T = f_T(U_T,X)$, $U_{\epsilon} = f_\epsilon(W)$, $Y = f_Y(X,T,U_{\epsilon})$. 
We further assume \emph{strong ignorability} (\emph{i.e.} no hidden confounders between $T$ and $Y$).

Using notations introduced in \citet{pearl2000models}, we define the potential response of a variable $Y$ to an action $do(T=t)$ for a particular realization of $U=u$ as $Y_t(u)$. Our goal is to predict the counterfactual response, for a new treatment assignment ($T=t'$), conditioned on an observed initial treatment response. That is, the probability of observing a different treatment response under treatment $t'$, after observing treatment response $y$ for covariate $x$ and treatment $t$. The probability density function of counterfactual $y'$ then writes:

\begin{align}
    p(Y_{t'} = y' \mid X=x, Y=y, T=t) &= \frac{p(Y_{t'}=y',X=x,T=t,Y=y)}{p(X=x,T=t,Y=y)} \nonumber\\
    &= \int_u p(Y_{t'}(u)=y') p(U = u\mid X=x, T=t, Y=y), \label{eq:counterfactuals}
\end{align}

and we refer to the counterfactual probability measure as $\nu_{t'}(x,y,t)$. Equation~\ref{eq:counterfactuals} suggests a natural three step procedure for computing the probability of counterfactual. First, the \emph{abduction} step infers the density of $U$ conditioned on the observed treatment outcomes, covariates and treatments: $p(U = u\mid X=x, T=t, Y=y)$. Second, in the \emph{action} step, one sets the new treatment in the causal model ($do(T=t')$). Lastly, in the \emph{prediction} step, one can propagate the values of $U=u$ and $T=t'$ in the causal graph, using $F$, to compute $p(Y_{t'}(u)=y')$.

In practice, we only have access to a set of $N$ observations of variables $X$, $Y$ and $T$. We refer to this dataset as $\mathcal{D} = (\mathbf{X},\mathbf{T},\mathbf{Y})$ where $\mathbf{X} = \{ x_i : i=1,...,N\}$, $\mathbf{Y} = \{ y_i : i=1,...,n\}$ and $\mathbf{T} = \{ t_i : i=1,...,N\}$. Importantly, we don't have access to counterfactual examples (\emph{i.e.} a tuple $(x,y,t,t',y')$), such that direclty learning a map $(x,y,t,t') \rightarrow y_{t'}$ is excluded.

\subsection{General Non-identifiability of Counterfactuals}

Because the background variables $U$ are hidden, the above three-steps procedure requires knowledge of the structural functions $F = \{f_X,f_T,f_Y\}$. Indeed, one can show that there exist multiple structural functions $F$ that would lead to the same observed joint density $p(X,Y,T)$ but would lead to incorrect counterfactual probabilities~\cite{pearl2009causal,pearl20217}. The correct causal model is thus in general non-identifiable, leading to non-identifiability of the counterfactual probability. We specify what is meant by the identifiability of counterfactuals in the following definition.

\begin{definition}[Identifiability of Counterfactuals]
Let $\rho$ be a metric on $\mathcal{P}(\mathcal{Y})$, $\nu_{t'}(X,Y,T)$ the true counterfactual probability measure and $\hat{\nu}_{t'}(X,Y,T)$ the estimator of the counterfactual probability measure with $N$ data points. Counterfactuals are $\rho$-identifiable at threshold $\delta$ if, for all $t,t'\in \mathcal{T}, x \in \mathcal{X}, y \in \mathcal{Y}$,

\begin{align*}
\lim_{N\rightarrow \infty} \rho(\nu_{t'}(x,y,y),\hat{\nu}_{t'}(x,y,t)) \leq \delta
\end{align*}

\label{def:identifiability}
\end{definition}

\vfill

\subsection{Causal Model Assumptions for Counterfactual Idenfiability}
\label{sec:assumptions}

Despite the general non-identifiability of structural equation models laid out above, we propose plausible assumptions that one can build upon to identify counterfactuals reliably. We first assume $X$ and $Y$ are continuous (potentially high dimensional) variables (such as images or time series). The treatment assignment $T$ is also assumed continuous, and $\mathcal{X} \times \mathcal{T}$ is a connected space. Our first central assumption posits that the hidden variable $U_{\epsilon}$ factorizes into a categorical and a continuous variable.

\begin{assumption}[Categorical Background Variable]
The background variable $U_{\epsilon}$ decomposes into a categorical latent variable $U_Z \in [K] = \{1,..,K\}$ and an independent exogenous continuous variable $U_{\eta}$.
\label{ass:categorical}
\end{assumption}

This assumption is depicted in the graphical model of Figure~\ref{fig:counterfactual_dag2} and embodies the intuition of different hidden groups that drive the treatment response. For instance, a treatment could have different responses depending on the stage of the disease a patient finds themself in. The disease stage is unobserved yet correlated with the observed covariates $X$ (through $W$). 

Due to the categorical nature of variable $U_Z$, one can write the conditional density of $Y$ as a mixture model:

\begin{align*}
p(Y=y \mid X = x, T = t) = \sum_{u_Z\in \{1,..,K\}} P(U_Z = u_Z) \cdot \int p(U_{\eta}=u_{\eta}) \mathbb{I}[f_Y(x,t,u_Z,u_{\eta})=y] du_{\eta}
\end{align*}

We define $\gamma$ as the probability density function of the conditional treatment response generated by $f_Y$, $U_Z$ and $U_{\eta}$. $\gamma$ is thus a mixture probability density function with mixture components\footnote{The mixture components are defined such that for any subset $\mathcal{A} \subset \mathcal{Y}$, we have $\int_{\mathcal{A}}{\gamma_k(X,T)(y)dy} = \int_{\infty}^{\infty} \mathbb{I}[f_Y(X,T,U_Z=k,U_{\eta}) \in \mathcal{A}] dP(U_{\eta})$. The mixutre weights are defined as $\alpha_k = P(U_Z=k)$.} $\gamma_k \in \mathcal{P}(\mathcal{Y})$ and weights $\omega_k$.

\begin{align}
    Y\mid X,T \sim \gamma(X,T) = \sum_{k=1}^{K} \omega_k \gamma_{k}(X,T)
    \label{eq:mixture}
\end{align}

 Without loss of generality, we assume that $U_\eta \sim \mathcal{N}(\mathbf{0},\Sigma^2)$. In the case of additive noise, the conditional distribution of Y becomes a mixture of Gaussians: $\gamma(X,T) = \sum_{k=1}^{K} \omega_k \mathcal{N}(\mu_{k}(X,T), \Sigma_k^2)$, where $\mu_k$ are functions mapping $X$ and $T$ to the mean of the mixture components and we consider different variances for each $k$. We now proceed with the next assumptions.

\begin{assumption}[Continuity]
The moments of the probability density functions $\gamma_k (x,t)$ exist and are continuous functions of $X$ and $T$:

\begin{align}
\mu^r_k(x,t) = \mathbb{E}_{Y \sim \gamma_k(x,t)}\left[Y^r\right] \in C(x,t) \quad \forall r \in \mathbb{N}, k \in [K]
\end{align}
\label{ass:continuity}
\end{assumption}

\begin{assumption}[Clusterability]
For each $(x,t)\in (\mathcal{X},\mathcal{T})$, the density $\gamma(x,t)$ is clusterable and the expected deviation of each $\gamma_k(x,t)$ is bounded by a constant $\delta \in \mathbb{R}$. That is, $\forall{k} \in [K] \text{ , } \forall x,t \in (\mathcal{X} \times \mathcal{T})$, with $\mu_k(x,t) = \mathbb{E}_{Y\sim \gamma_k(x,t)}[Y]$:
\begin{align}
    \mathbb{E}_{Y \sim \gamma_k(x,t)} \big[ \lVert Y - \mu_k(x,t) \rVert_2 \big] \leq \delta
\end{align}
\label{ass:clusterability}
\end{assumption}

In our motivating clinical example, Assumption \ref{ass:continuity} reflects that the probability of specific treatment response changes continuously over the set of observed covariates and treatments. In particular, the expected treatment outcome for a particular patient varies continuously with the treatment assignment, which is a common assumption, \emph{e.g.} in clinical practice~\cite{monnet2016prediction}. 

Assumption \ref{ass:clusterability} posits \emph{clusterability} of the mixture components $\gamma_k$ for which a rigorous mathematical definition is given in Appendix \ref{app:proofs}. It is motivated by recent results on the identifiability of mixtures 
models\cite{aragam2020identifiability}. Intuitively, it supposes that patients with the same observed covariates and treatment assignment but different hidden group will show different treatment outcomes. We also bound the expected deviation of the mixture components $\gamma_k$ that characterize the inter-group variability in the treatment response for a particular patient and treatment outcome.



\section{Methods}

\subsection{Identifiability and Counterfactuals Reconstruction}

For a fixed point $(X = x, T = t)$, Equation \ref{eq:mixture} is a finite mixture model, for which identifiability results are available \cite{aragam2020identifiability}. Notably, these results guarantee identifiability up to a permutation of the latent class assignment $\sigma(\cdot) : [K] \rightarrow [K]$. That is, there exists some permutation $\sigma(\cdot) : [K] \rightarrow [K]$ such that $\hat{\gamma}_{\sigma_k(k)(x,t)} \approx \gamma_k(x,t)$, $\hat{\omega}_{\sigma_k(k)} \approx \omega_k$, where $\hat{\gamma}$ and $\hat{\omega}$ are the estimated density functions and weights. However, it does not entail identifiability of the counterfactuals in the sense of definition~\ref{def:identifiability}. Indeed, the action step of the counterfactual strategy from Section \ref{sec:problem_setup} requires a consistent permutation $\sigma$ across the whole domain $(\mathcal{X}\times\mathcal{T})$ in order to reuse the inferred class assignments $\hat{U}_Z$ at a specific  point $(X = x,T = t)$ to predict the counterfactual at another point $(X = x,T = t')$ --- with a different treatment assignment. Nevertheless, using the assumptions from the previous section, we can still ensure the identifiability of the counterfactuals as the following result confirms:

\begin{theorem}[Identifiabilty of Counterfactuals with Categorical Background Variables]
Let $X$, $T$ and $Y$ be continuous random variables generated according to the graphical model of Figure\ref{fig:counterfactual_dag2} with the domain of $X$ and $T$ being connected. Let $W_1(\cdot,\cdot)$ be the first Wasserstein distance on $\mathcal{P}(\mathcal{Y})$, $\nu_{t'}(X,Y,T)$ the probability distribution of $Y_{t'}\mid X,Y,T$ and $\hat{\nu}_{t'}^N(X,Y,T)$ its estimator from $N$ observed data points. 
If Assumptions \ref{ass:categorical}, \ref{ass:continuity} and \ref{ass:clusterability} hold, for each $(x,t)$, the counterfactual distribution is $W_1$-identifiable in expectation at threshold $\delta$:

\begin{align*}
    \lim_{N\rightarrow \infty} \mathbb{E}_{Y\sim\gamma(x,t)}\left[W_1(\nu_{t'}(x,Y,t),\hat{\nu}_{t'}^N(x,Y,t) )\right] \leq \delta
\end{align*}

In the special case when the noise response is additive, we have

\begin{align*}
    \lim_{N\rightarrow \infty} W_1(\nu_{t'}(X,Y,T),\hat{\nu}_{t'}^N(X,Y,T)) = 0
\end{align*}

\label{result}
\end{theorem}

The proof is given in Appendix \ref{app:proofs}. This result gives us a bound on the distance between the inferred and true counterfactual distributions in the asymptotic regime. Importantly, it does not restrict the dimension of $\mathcal{X}$ and $\mathcal{Y}$, and is thus valid on challenging data modalities such as time series or images.

\paragraph{Continuity of distribution and complexity}

The result above holds asymptotically in the number of available samples. In the additive Gaussian case, the sample complexity for learning a $K$-mixture model with $Y\in \mathbb{R}^d$ within $\epsilon$ total variation distance is $\Tilde{O}(Kd^2/\epsilon)$~\cite{ashtiani2018nearly}. Fortunately, the continuity assumption (Assumption \ref{ass:continuity}) saves us from having to learn an individual mixture at each point $(X=x,T=t)$, by jointly learning the continuous moments functions $\mu'_r(X,T)$. A better sample complexity bound can then be derived with further assumptions on $\mu'_r(X,T)$.



\subsection{\method~: CounterFactual Query Prediction}
\label{sec:cfqp}

\begin{figure}
    \centering
    \includegraphics[width=\textwidth]{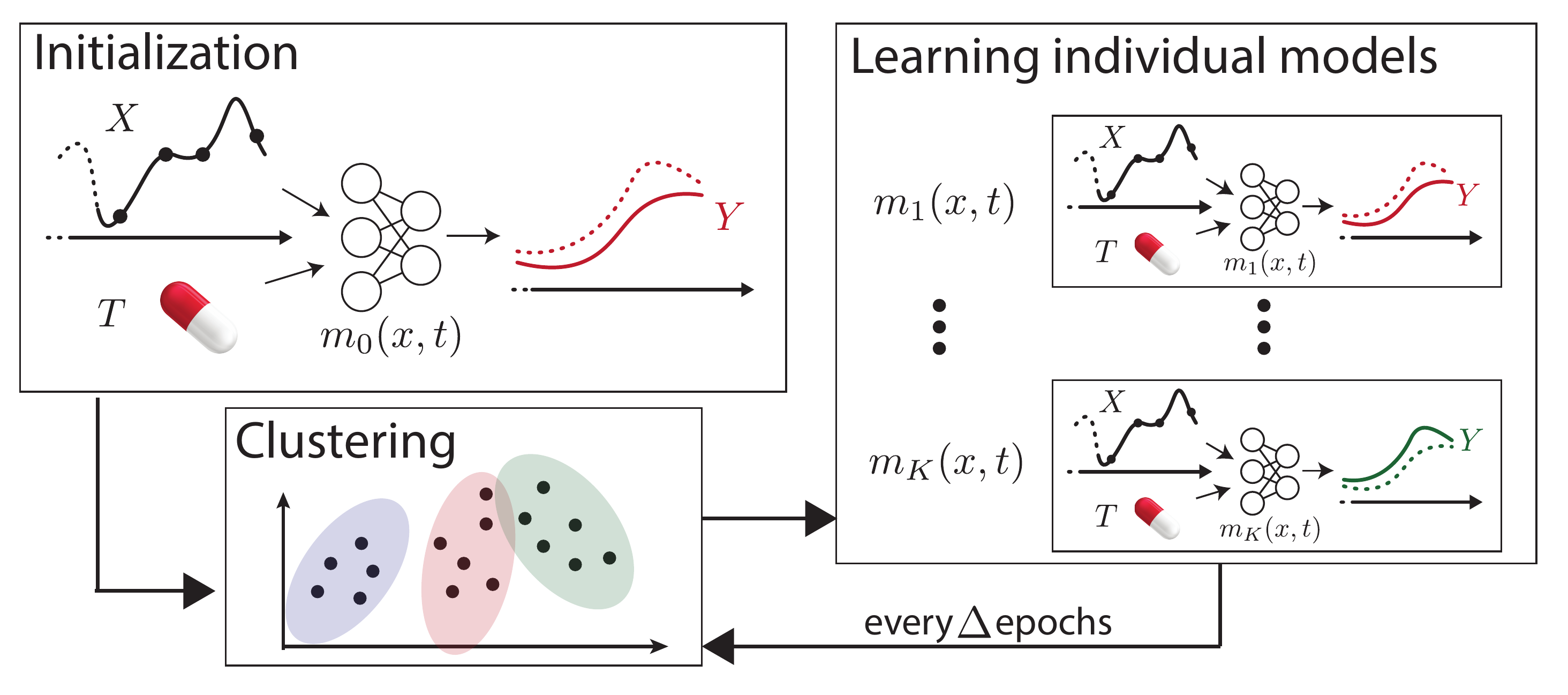}
    \caption{Training procedure of \method. We first initialize a common model $m_0$. Second, we cluster the different available data points into $K$ clusters. We then use this clustering to train $K$ individual models (Maximization step). The clustering is updated every $\Delta$ epochs (Expectation step).}
    \label{fig:model_architecture}
\end{figure}

Equipped with those theoretical results, we introduce \method, a counterfactual prediction model based on a neural Expectation-Maximization mechanism. The basic building block of \method~is a base-model $m(x,t)$, that predicts the treatment response $y$ based on covariates and treatment assignment. For each latent category $k$, we learn a base-model that approximates the individual treatment response in that category : $m_k(x,t) \approx \mu_k(x,t)$. Our theoretical results require the true number of classes of $K_0$ to be known in advance. Yet, this is rarely the case in practice, and we describe our architecture for an arbitrary number of classes $K$. 
The learning of the base-models follows three steps: a joint initialization, an expectation phase, and a maximization phase. The overall process is depicted in Figure \ref{fig:model_architecture}. We also present a pseudo-code description of the procedure in Algorithm \ref{alg:cf_reconstruction}. 

\paragraph{Initialization} We first train a single common base-model $m_0$ on all data points $\mathbf{X},\mathbf{T}$ and $\mathbf{Y}$. Because it discards the variability of the hidden variable $U_Z$, this model will approximate the conditional average treatment response : $m_0(x,t) \approx \mathbb{E}[Y\mid X=x,T=t]$.

\paragraph{Expectation - Clustering} 
The clustering step ensures that each base-model $m_k$ is trained with the data from the corresponding latent category. We distinguish between an \emph{initial} clustering and an \emph{update} clustering stage.

The \emph{initial} clustering assignment happens after the initialization step has converged. We then use the clusterability assumption (Assumption \ref{ass:clusterability}) to drive the assignment and use $K$-means clustering based on the residuals $\lVert y-m_0(x,t) \rVert_2^2$.

The \emph{update} clustering step happens at regular intervals to improve the quality of the cluster assignment as the performance of each individual base-model increases. A point $(x,y,t)$ is assigned to cluster $k = \underset{j}{\arg\min} \lVert y-m_j(x,t) \rVert_2^2$. 

\paragraph{Maximization - Learning individual base-models}

Once the dataset $\mathcal{D} = (\mathbf{X},\mathbf{T},\mathbf{Y})$ is clustered in $K$ groups, we train the individual models $m_k$ on the corresponding cluster. This corresponds to the maximization step of the expectation-maximization scheme. Every $\Delta$ epochs, we update the cluster assignment until convergence.

\paragraph{Counterfactual prediction}

For a data point $(x,t,y)$, we first infer the cluster assignment $k = \underset{j}{\arg\min} \lVert y-m_j(x,t) \rVert_2^2$. We predict the counterfactual for treatment $t'$ as $\hat{y'} = m_k(x,t')$.


\begin{algorithm}
\caption{\method~Training}\label{alg:cf_reconstruction}
\KwData{$\mathbf{X}$,$\mathbf{Y}$,$\mathbf{T}$, the number of latent clusters $K$, a number of epochs $e_{max}$, the update-period $\Delta$.}
\KwResult{A list of $K$ hidden models $m_k$.}
Initialize a single base model $m_0$ at random.\\
\For{epoch$\ \gets0$ \KwTo $e_{max}$}{
Train $m_0$ minimizing $\mathcal{L}_0 = \mathbb{E}\left[(m_0(X,T)- Y)^2\right]$\\
}
Compute residuals $\mathbf{r} = (m_0(\mathbf{X},\mathbf{T})- \mathbf{Y})$.\\
Assign the residuals of each training sample into $K'$ clusters (initial).\\
Initialize each $m_k$ with $m_0$.\\
\For{epoch$\ \gets0$ \KwTo $e_{max}$}{
  \For{$i\gets0$ \KwTo $\Delta$}{
    Train $m_k$ minimizing $\mathcal{L}_k = \mathbb{E}\left[(m_k(X_k,T_k)- Y_k)^2\right]$ for each cluster.\\
    }
Compute residuals $\mathcal{R} = \{(m_k\left(\mathbf{X},\mathbf{T})- \mathbf{Y}\right) :   k = 1,...,K\}$.\\
Assign the residuals of each training sample into $K$ clusters (update).\\
}
\textbf{Return} trained models $m_k$.
\end{algorithm}

\section{Related Work}
Causal perspectives in machine learning have gained significant traction in the past years \cite{scholkopf2021toward,scholkopf2022causality}. Among them, one distinguishes between causal \emph{discovery} approaches, aiming at discovering the causal relations between variables \cite{glymour2019review,de2020latent}, and causal \emph{inference}, aiming at building treatment effects estimators from data \cite{pearl2003statistics}. Our work belongs to the latter. A common way to estimate counterfactuals is to posit a specific structural causal model~\cite{pearl20217} or its specific functional form~\cite{balke2013counterfactuals}, therefore ensuring identifiability. Synthetic controls are an example of this strategy, assuming an underlying linear structure \cite{abadie2021using}. However, these approaches are by definition restrictive in terms of the expressivity of the structural equations. Deep learning approaches for counterfactual estimation have been recently proposed to address this issue \cite{pawlowski2020deep,sanchez2021diffusion} but without identifiability guarantee. For the discrete case, assumptions have been proposed to bridge this gap, such as monotonicity or generalization thereof~\cite{oberst2019counterfactual,lorberbom2021learning}. Nevertheless, because they focus on addressing the discrete case, they are not directly applicable to high-dimensional data such as time series, which is a motivation for our work.

Our work builds upon the literature on the identifiability of finite mixture distributions where guarantees have been achieved in the Gaussian \cite{dasgupta1999learning} and non-parametric cases \cite{aragam2020identifiability}, among others.

\section{Experiments}
\label{sec:experiments}

\subsection{Datasets}

Treatment outcomes are often complex and high-dimensional. We thus evaluate our proposed model architecture on high-dimensional data modalities: time series and images. 
As counterfactual inference evaluation on real-world data is a complex and ongoing research area, we use synthetic datasets inspired by case studies from the literature. Details about the data generation are given in Appendix~\ref{app:data_generation}.

\subsubsection{Counterfactual Image Transformation}
To explore the capacity of our model to operate on images, we generate a dataset inspired by \cite{pawlowski2020deep}, building upon modification of MNIST images. Covariates $X$ are original MNIST images and the treatment $T$ is a spatial rotation applied to the image. $U_Z$ is a coloring of the image and the treatment outcome $Y$ is the resulting colored and rotated MNIST image. $U_{\eta}$ is either an additive Gaussian noise at each pixel or a Gaussian blur with a random kernel.

\subsubsection{Counterfactual Longitudinal Treatment Effect Prediction}

Next, we explore the performance of our model on time series data. As per our motivational example, our goal is here to infer counterfactuals from clinical trajectories with a hidden patient class. $X$ and $Y$ are multi-dimensional time series, $U_Z$ is a latent patient group and $T$ is a continuous treatment assignment. We used two different datasets: an harmonic oscillator and a cardiovascular model.

\paragraph{Harmonic oscillator.}
We simulate the angular positions of two coupled harmonic oscillators. The treatment consists of applying different gradual offsets to the time series. We consider three hidden groups ($K_0=3$) that modulate the treatment response differently.

\paragraph{Cardiovascular simulator.}

We use a simulator of the cardiovascular system proposed in \cite{zenker2007inverse} to predict the impact of fluid intake on blood pressure. Fluids are commonly administered in intensive care for treating severe hypotension. However, the individual patient response is difficult to assess a priori, as it depends on clinically hidden variables. We consider here two hidden groups of patients with distinct responses to fluid intake. $X$ and $Y$ contain the arterial and venous blood pressure time series before and after fluid intake. The treatment is the amount of fluid injected. We add process noise by either considering an additive Gaussian noise on the observed treatment response or by introducing noise in the fluid injection process, leading to a non-linear and complex perturbation.

\vfill

\subsection{Baselines}

We compare our approach against multiple counterfactual inference models from the literature: \emph{Diff-SCM}~\cite{sanchez2021diffusion}, a recently introduced diffusion based counterfactual model; \emph{Deep-SCM}~\cite{pawlowski2020deep}, a deep counterfactual inference architecture building on normalizing flows ; Synthetic Controls (\emph{SC})~\cite{abadie2021using}, a well known linear method for counterfactuals estimation and \emph{Deep-ITE}, a deep individual treatment effect estimator such as proposed in \cite{shalit2017estimating,johansson2016learning}, using direct predictions from $X$ and $T$.

\begin{table}[htbp!]
\centering
\caption{Test MSE of the counterfactual reconstructions for the different datasets.}
\begin{adjustbox}{width=\textwidth}
\begin{tabular}{lccc|ccc}
\toprule[1.5pt]
  & \multicolumn{3}{c}{Additive Noise} & \multicolumn{3}{c}{Non-Additive Noise} \\
  \midrule
Model &  Harmonic Oscillator & colored-MNIST & Cardiovascular & Harmonic Oscillator & colored-MNIST & Cardiovascular\\
\midrule[1.5pt]
Deep-ITE~\cite{johansson2016learning} & $0.187 \pm 0.006$ & $0.017 \pm 0.001$ & $ 1.084\pm0.087 $ & $ 0.174\pm0.004 $ & $ 0.017\pm0.001 $ & $ 1.14\pm0.121 $\\
\midrule
SC~\cite{abadie2021using} & $0.177 \pm 0.131$ & $ 0.020\pm0.001 $ & $ 1.610\pm0.141 $ & $ 0.167\pm0.004 $ & $ 0.020\pm0.001 $ & $ 1.628\pm0.144 $\\
\midrule
Deep-SCM~\cite{pawlowski2020deep} & $0.124 \pm 0.005$ & $0.011 \pm 0.001$  &  $0.405 \pm 0.042$ & $0.123 \pm 0.005$ & $0.011 \pm 0.001$ & $0.424 \pm 0.042$\\
\midrule
Diff-SCM~\cite{sanchez2021diffusion} & $0.082 \pm 0.023$ & $0.008 \pm 0.004$ & $ 0.206\pm0.036 $ & $ 0.106\pm0.038 $ & $ 0.009\pm0.002 $ & $ 0.311\pm0.073 $ \\
\midrule
\method~(ours) & $ \mathbf{0.013 \pm 0.001}$ & $\mathbf{0.001 \pm 0.001}$ & $\mathbf{0.077 \pm 0.050}$ & $ \mathbf{0.009\pm0.001} $ & $ \mathbf{0.002\pm0.001} $ & $ \mathbf{0.188\pm0.114} $\\
\bottomrule
\end{tabular}
\end{adjustbox}
\label{tab:CF_rec}
\end{table}

\subsection{Counterfactuals Prediction}

\begin{wrapfigure}{R}{0.48\textwidth}
\begin{center}
\includegraphics[width=\linewidth]{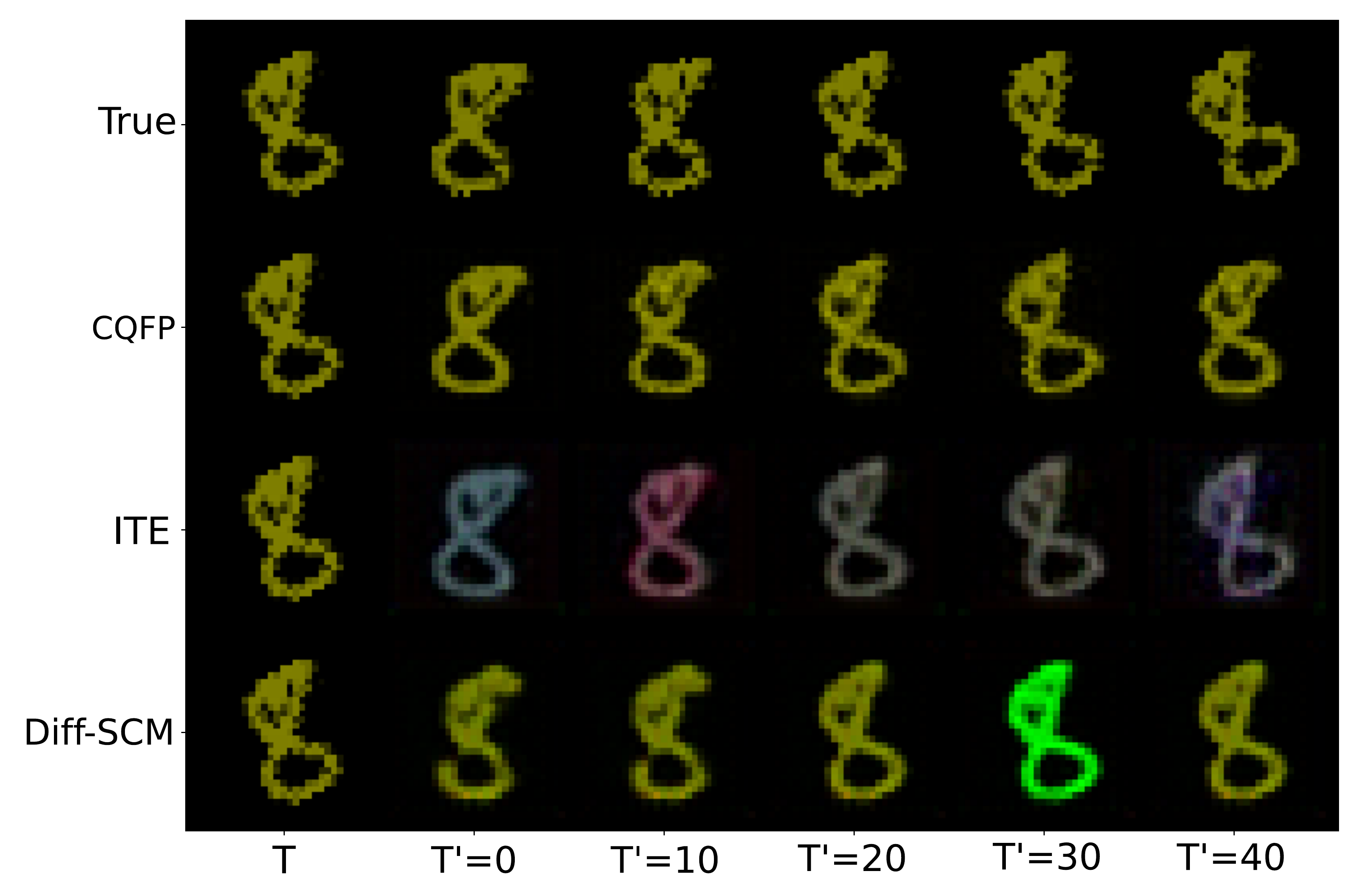}
\end{center}
\caption{Example of reconstructed counterfactuals on the image dataset for different methods (best seen in color). Here, $U_Z$ corresponds to the color of the digit and $T$ to the rotation angle. $X$ is the non-rotated and non-colored image while $Y_T$ is the colored and rotated image. First column is the observed factual treatment outcome. Subsequent columns show the counterfactual reconstructions for different values of $T'$. First row is the ground truth.}
\label{fig:cf_mnist_exps}
\end{wrapfigure}

We evaluate our approach in terms of the quality of the counterfactual reconstructions. We generate counterfactuals ($t'$,$y'$) for each observation ($x$,$y$,$t$) in the test set, resulting in a tuple ($x$,$y$,$t$,$t'$,$y'$). We evaluate the MSE between the true counterfactual $y'$ and the estimated one $\hat{y}'$ as $\mbox{MSE}_{cf} = \lVert y' - \hat{y}'\rVert_2^2$. We set the number of clusters $K$ as an hyper-parameter and select the value that result in lowest validation error. The results are presented in Table~\ref{tab:CF_rec}. We use additive and non-additive noise variants for each dataset. For each dataset and each variant, we see that our approach outperforms the other baselines, demonstrating its experimental effectiveness.

In Figure \ref{fig:cf_mnist_exps}, we also report examples of counterfactual reconstructions on the image dataset. Each row corresponds to counterfactual estimations for a particular method (top is the truth), and each column represents a different treatment assignment $T'$ (the left column is the factual $T$). Because it cannot model the categorical latent variable, the Deep-ITE model gets the expected individual treatment effect right but fails to accurately estimate the counterfactuals. For Diff-SCM, we observe that, despite providing an accurate reconstruction for most of the treatment assignments, the lack of guarantees on the deep structural equation model leads to potential incorrect reconstructions (\emph{e.g.} at $T'=30$). To assess the quality of the counterfactual images reconstructions, we also evaluate the structural similarity index and report the results in Appendix \ref{app:SSIM}. We observe that \method~outperforms all baselines on that metric.

\subsection{Robustness Analysis}
\label{sec:robustness}
We further complement our experimental section with a comprehensive robustness analysis. 
We first investigate the impact of misspecifying the number of latent classes (\emph{i.e.} when $K_0 \neq K$). We then investigate the impact of a correlation between $X$ and $Z$ with increasing strength. Finally, we investigate the robustness of our approach to different noise responses.

\subsubsection{Number of latent classes}

We study the evolution of performance when the number of classes $K$ is misspecified ($K_0 \neq K$). In Figure~\ref{fig:num_groups}, we show the reconstruction error on the validation set (\emph{i.e.}, on the factual data) and the counterfactual reconstruction error of \method~for different values of $K$ on the image dataset with true number of latent classes $K_0 = 6$. The validation MSE is lowest for $K=6$ and $K=7$, which corresponds to optimal test reconstruction error. This hyper-parameter can thus easily be tuned by monitoring the treatment prediction performance on the factual data. In Appendix \ref{app:number_groups}, we further report the quantitative results for the other datasets in function of the number of the clusters $K$.

\subsubsection{Correlation between $X$ and $U_Z$ }
\label{sec:correlation_exp}
\begin{figure}[htbp]
    \centering
    \begin{subfigure}[t]{0.45\textwidth}
    \centering
    \includegraphics[width=1.1\linewidth]{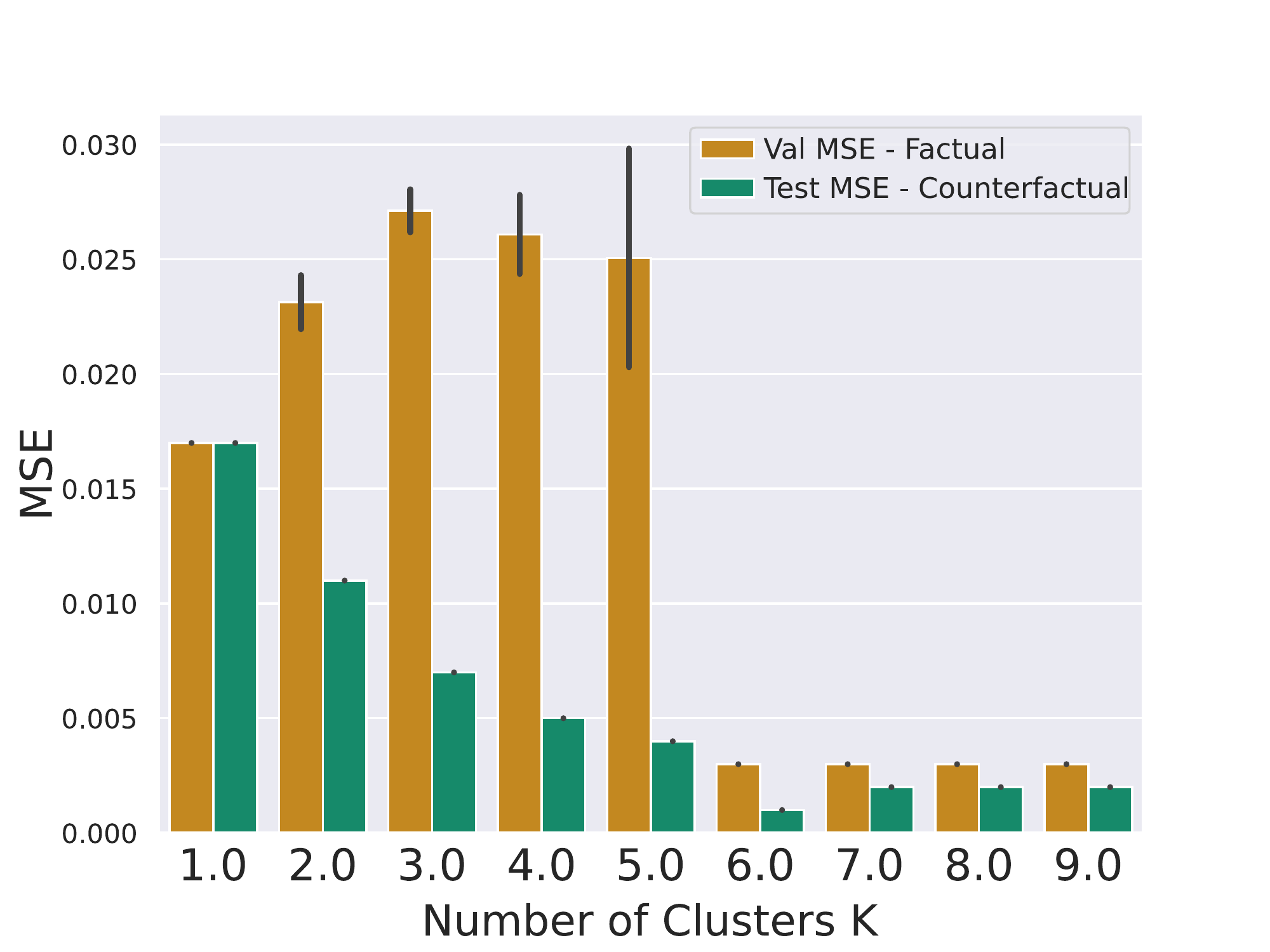}
    \caption{Reconstruction MSE in function of the number of groups. The true number of groups is 6.}
    \label{fig:num_groups}
    \end{subfigure}
    \hfill
    \begin{subfigure}[t]{0.5\textwidth}
    \centering
    \includegraphics[width=\linewidth]{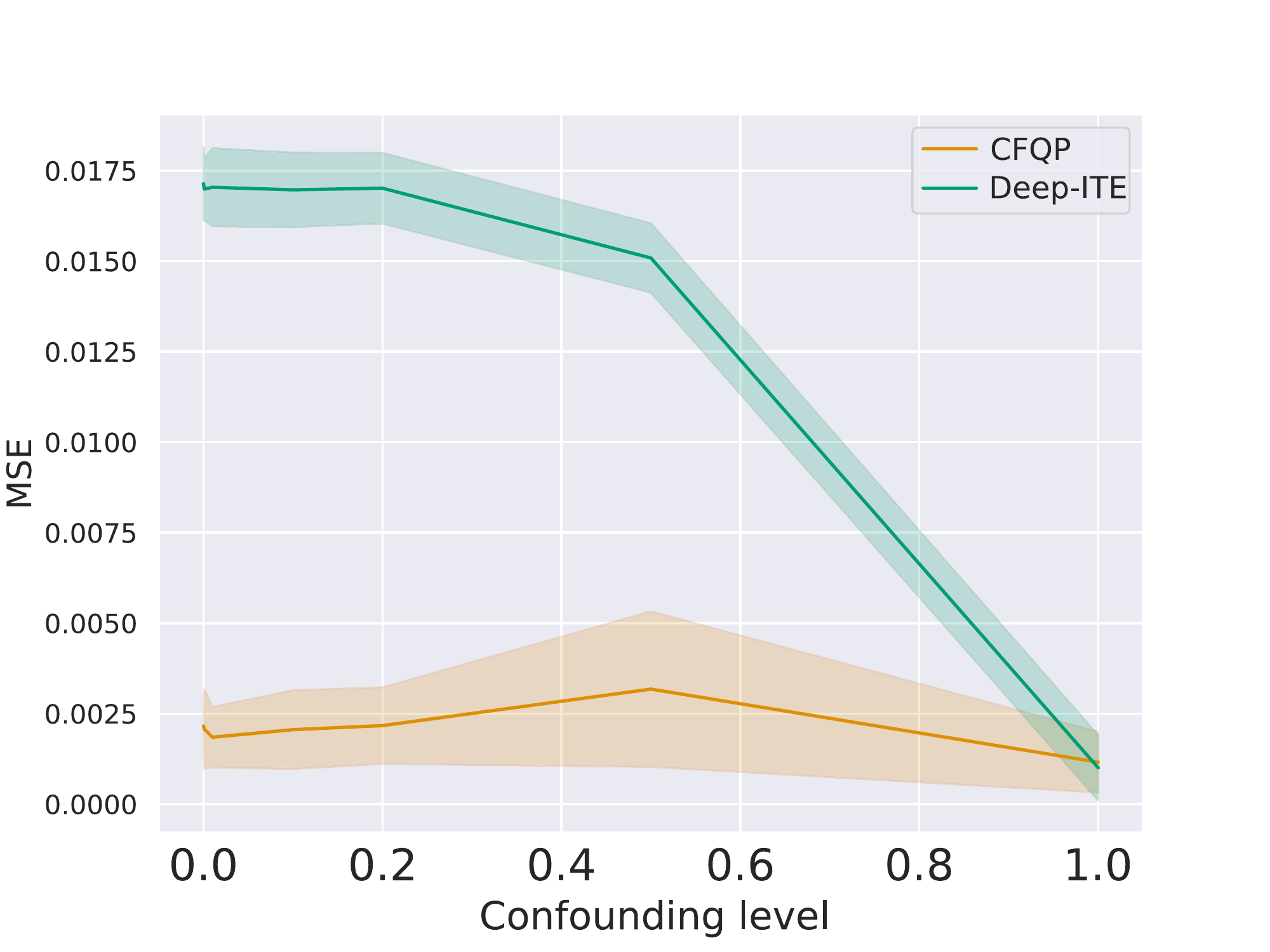}
    \caption{Evolution of the reconstruction MSE in function of the correlation strength between $X$ and $U_Z$.}
    \label{fig:correlation}
\end{subfigure}
\hfill
\caption{Robustness analysis for the number of latent classes (left) and correlation strength (right).}
\end{figure}

Our approach holds even when the latent background variable $U_Z$ and the covariates $X$ are correlated. When this is the case, information from $X$ can be used to infer the value of $U_Z$. In Figure \ref{fig:correlation}, we study the impact of the correlation strength $\rho$ between $X$ and $U_Z$. More details about the definition of $\rho$ are to be found in Appendix \ref{app:experiments}. We compare the performance of~\method~with a Deep-ITE model. We observe that the counterfactual performance of our model remains constant, regardless of the correlation strength. However, as the correlation becomes more important, $U_Z$ becomes a deterministic outcome of $X$ and Deep-ITE eventually converges to the performance of \method.

\subsubsection{Non-additive Noise Responses}

\begin{wraptable}{R}{0.35\textwidth}
\vspace{-20pt}
\centering
\caption{Fluids $\mbox{PEHE}$}
\begin{adjustbox}{width=0.35\textwidth}
\begin{tabular}{lc}
\toprule[1.5pt]
Model & Cardiovascular $\mbox{PEHE}$ $\downarrow$ \\
\midrule[1.5pt]
Deep-ITE~\cite{johansson2016learning} &  $ 0.594\pm0.160 $\\
\midrule
SC~\cite{abadie2021using} & $ 0.258\pm0.016 $ \\
\midrule
Diff-SCM~\cite{sanchez2021diffusion} &  $ 0.377\pm0.049 $ \\
\midrule
\method~(ours) & $\mathbf{0.143\pm0.108}$ \\
\bottomrule
\end{tabular}
\end{adjustbox}
\label{tab:ITE}
\vspace{-20pt}
\end{wraptable}

Our model architecture is motivated by the additive noise case. Yet, our theoretical results hold for arbitrary distributions. In Table \ref{tab:CF_rec}, we  reports the performance of the different models under non-additive noise responses. We observe an overall small degradation of performance of all methods but \method~still outperforms baselines, as predicted by our identifiability results.

\subsection{Retrospective Individualized Treatment Effect Estimation}

We explore the ability of our model to infer individualized treatment effects retrospectively. Based on observed treatment outcomes, we predict the difference in treatment response between two treatment regimes for a single individual. The metric of interest is the Precision in Estimation of Heterogeneous Effects $\mbox{PEHE}$ \cite{hill2011bayesian}.
For two treatment regimes $t'$ and $t''$, we write 
\begin{align}
\mbox{PEHE}(t',t'') = \sqrt{\mathbb{E}_{X,T,Y} \left[ \left((Y_{t''}-Y_{t'})-(\hat{Y}_{t''}-\hat{Y}_{t'})\right)^2 \mid X,Y,T \right]}.
\end{align}
In Table~\ref{tab:ITE}, we report $\mbox{PEHE}(t'=0.5,t''=0.8)$ for test patients in the Cardiovascular dataset with different fluid intake. We observe that the methods leveraging information about treatment outcomes, \method, SC and Diff-SCM outperform the classical individual treatment effect estimators (Deep-ITE).

\subsubsection{Impact of the clustering algorithm}

Our approach relies on an \emph{initial} clustering step as described in Section \ref{sec:cfqp}. Different clustering strategies can be considered for this step. In Appendix \ref{app:clustering}, we investigate the difference between K-means and Gaussian mixture models. Only minor deviations are observed in terms of performance between both approaches.

\section{Conclusion}

Estimating counterfactuals from observational studies is one of the most challenging tasks in causal inference. In this work, we proposed a set of reasonable assumptions that allow computing counterfactuals on high-dimensional data while harnessing the power of modern machine learning architectures. Based on these assumptions, we derived a new counterfactual model and demonstrated favorable experimental performance. In particular, we showed that this approach could be used to infer individual treatment effects a posteriori in clinical patient trajectories. Nevertheless, the set of assumptions proposed here is not unique, and depending on the specific applications, others might be deemed more relevant. Indeed, the space of assumptions that allow to approximately recover counterfactuals with a controlled level of error is a potentially  very rich research direction. Other model architectures compatible with our set of assumptions are also possible.

\paragraph{Potential negative societal impacts} Counterfactual inference has been used in clinical settings and for assessing the impact of public policies~\cite{abadie2010synthetic}, among others. However, as shown in this paper, its correctness hinges on assumptions whose validity is not always met, potentially leading to misleading conclusions.

\paragraph{Acknowledgements} Edward is funded by a FWO-SB grant. Edward would also like to thank Zeshan Hussain, David Sontag and Adam Arany for their comments on preliminary versions of this manuscript. 

\clearpage

\bibliographystyle{abbrvnat}
\bibliography{cf}

\clearpage

\section*{Checklist}

\begin{enumerate}

\item For all authors...
\begin{enumerate}
  \item Do the main claims made in the abstract and introduction accurately reflect the paper's contributions and scope?
    \answerYes{}
  \item Did you describe the limitations of your work?
    \answerYes{The assumptions are cleary defined in section \ref{sec:assumptions}. The conclusion also relfects on the limitations of our work.}
  \item Did you discuss any potential negative societal impacts of your work?
    \answerYes{Last paragraph in the conclusion discusses the potential negative societal impacts of counterfactual inference. The main risk we identify is blind acceptance of the assumptions required for correct inference, which can lead to misleading conclusions in assessing treatment effects as posteriori.}
  \item Have you read the ethics review guidelines and ensured that your paper conforms to them?
    \answerYes{ Our paper conforms to the ethics review guidlines.}
\end{enumerate}

\item If you are including theoretical results...
\begin{enumerate}
  \item Did you state the full set of assumptions of all theoretical results?
    \answerYes{The full set of assumptions is given in Section \ref{sec:assumptions}}
        \item Did you include complete proofs of all theoretical results?
    \answerYes{The proofs of our results are given in Section \ref{app:proofs} of the Appendix.}
\end{enumerate}

\item If you ran experiments...
\begin{enumerate}
  \item Did you include the code, data, and instructions needed to reproduce the main experimental results (either in the supplemental material or as a URL)?
    \answerYes{ The code is available at the following url: \url{https://anonymous.4open.science/r/cfqp}}
  \item Did you specify all the training details (e.g., data splits, hyperparameters, how they were chosen)?
    \answerYes{We provide the training details in Section \ref{app:experiments} of the Appendix}
        \item Did you report error bars (e.g., with respect to the random seed after running experiments multiple times)?
    \answerYes{Error bars are reported from running the experiments 5 times (5 folds)}
        \item Did you include the total amount of compute and the type of resources used (e.g., type of GPUs, internal cluster, or cloud provider)?
    \answerYes{The total amount of compute and the type of resources used is available in the Appendix Section \ref{app:experiments}}
\end{enumerate}

\item If you are using existing assets (e.g., code, data, models) or curating/releasing new assets...
\begin{enumerate}
  \item If your work uses existing assets, did you cite the creators?
    \answerYes{The authors of the code bases used for implementing the baselines are credited both in the code base and cited in the text. The list of software packages used is available at \url{https://anonymous.4open.science/r/cfqp/licenses.txt}}
  \item Did you mention the license of the assets?
    \answerYes{The licenses of all software packages used in our experiments is available at \url{https://anonymous.4open.science/r/cfqp/licenses.txt}. }
  \item Did you include any new assets either in the supplemental material or as a URL?
    \answerYes{We include an anonymized repository containing the code used for the experiments. It is available at \url{https://anonymous.4open.science/r/cfqp}}
  \item Did you discuss whether and how consent was obtained from people whose data you're using/curating?
    \answerNA{We are not using any personal data.}
  \item Did you discuss whether the data you are using/curating contains personally identifiable information or offensive content?
    \answerNA{None of our datasets contains personal data.}
\end{enumerate}

\item If you used crowdsourcing or conducted research with human subjects...
\begin{enumerate}
  \item Did you include the full text of instructions given to participants and screenshots, if applicable?
    \answerNA{No research with human subjects}
  \item Did you describe any potential participant risks, with links to Institutional Review Board (IRB) approvals, if applicable?
    \answerNA{No research with human subjects}
  \item Did you include the estimated hourly wage paid to participants and the total amount spent on participant compensation?
    \answerNA{No research with human subjects}
\end{enumerate}

\end{enumerate}


\clearpage

\appendix

\section{Datasets}
\label{app:data_generation}
\subsection{Image Dataset Generation}

The image dataset uses the MNIST handwritten digits images dataset. We use the following data generating process:

\begin{itemize}
    \item $X \in \{0,...,255\}^{28\times 28\times 1}$ is an MNIST image sampled at random from the dataset. We write $y_{label}$ the label of the digit present in the image.
    \item $T \in \mathcal{R} = \mathcal{U}(0,0.3) + 5\cdot \sigma\left(\frac{\frac{1}{784} \sum_{i=1}^{28} \sum_{j=1}^{28} X_{i,j}- 33}{11}\right)$ with $\sigma(\cdot)$ the sigmoid function. This encodes confounding between the observables $X$ and the treatment assignment $T$.
    \item $p_0 = \frac{1}{K}$
    \item $p \in [0,1]^{K} $
    \item $\rho \in [0,1]$ is the strength of confounding between $X$ and $U_Z$
    \item $p[i] = \frac{1-((1-p_0)\rho+p_0)}{K-1} \quad \forall i\in [K], i \neq y_{label} \mbox{mod} K$
    \item $p[i] = ((1-p_0)\rho+p_0) \quad \forall i\in [K], i = y_{label} \mbox{mod}(K)$
    \item $U_Z \sim \mathcal{M}\text{ultinomial}(p)$
    \item $Y \in \{0,...,255\}^{28\times 28\times 1}$ is the rotated image $X$ with angle $T$ and Gaussian blur with kernel size $5\times$ and standard deviation $\sigma$.
\end{itemize}

\begin{figure}[htbp]
    \centering
    \includegraphics[width=0.8\linewidth]{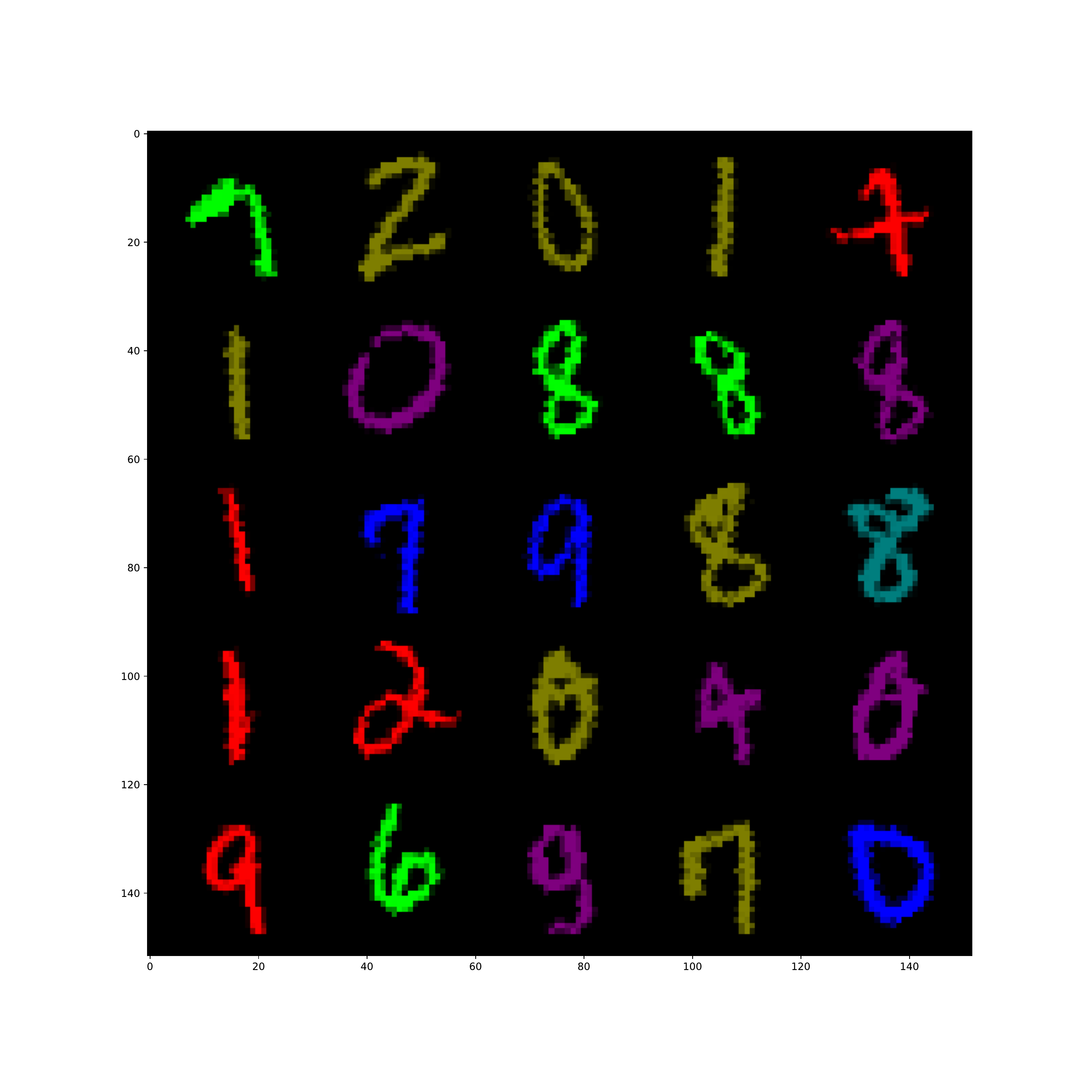}
    \caption{Examples of colored and rotated MNIST images $Y$}
    \label{fig:mnist_samples}
\end{figure}

Examples of generated images $Y$ are shown in Figure \ref{fig:mnist_samples}. We use the original MNIST training set that we randomly divide as $70\%$ training set, $15\%$ validation set and $15\%$ test set.

\subsubsection{Evaluating the structural similarity index measure of the counterfactual images predictions}
\label{app:SSIM}

Mean square error between the ground truth and predictions might not always be the most meaningful way to assess the quality of the reconstruction. A widely used metric in computer vision is the structural similarity index measure (SSIM). In Table \ref{tab:SSIM_MNIST}, we report the SSIM between the predicted and ground truth images counterfactuals. We observe that \method outperforms all baselines in that metric, producing more accurate counterfactual predictions.

\begin{table}[ht!]
\centering
\caption{Structural similarity results for the MNIST reconstructions}
\begin{adjustbox}{width=0.6\textwidth}
\begin{tabular}{lc|c}
\toprule[1.5pt]
  & \multicolumn{1}{c}{Additive Noise} & \multicolumn{1}{c}{Non-Additive Noise} \\
  \midrule
Model &  colored-MNIST  & colored-MNIST\\
\midrule[1.5pt]
Deep-ITE~\cite{johansson2016learning} & $ 0.490\pm0.006 $ & $ 0.484\pm0.004 $ \\
\midrule
SC~\cite{abadie2021using} & $ 0.020\pm0.001 $ & $ 0.020\pm0.001 $ \\
\midrule
Deep-SCM~\cite{pawlowski2020deep} & $0.857 \pm 0.003$ & $0.856 \pm 0.004$ \\
\midrule
Diff-SCM~\cite{sanchez2021diffusion} & $ 0.780\pm0.063 $ & $0.750 \pm 0.052$  \\
\midrule
\method~(ours) & $0.964 \pm 0.003$ & $0.959 \pm 0.005$ \\
\bottomrule
\end{tabular}
\end{adjustbox}
\label{tab:SSIM_MNIST}
\end{table}

\subsection{Harmonic Oscillator}

We consider input times series $X \in \mathbb{R}^{d_x\times t_x}$, with $d_x=2$ the number of temporal dimensions and $t_x$ the length of the input time series. The temporal responses are given by $Y \in \mathbb{R}^{d_y\times t_y}$. We use the following data generating process :


\begin{align*}
    U_Z &\sim \{0,1,2\} \\
     T &\sim \mathcal{U}(0.2,1) \\
     t_x &= (0,...,19) \\
     t_y &= (20,...,40)\\
      \phi &\sim \mathcal{N}(0,1) \\
    U_{\eta_x} &\sim \mathcal{N}(0,\sigma^2) \in \mathbb{R}^{t_x,2}\\
     U_{\eta_{y}} &\sim \mathcal{N}(0,\sigma^2) \in \mathbb{R}^{t_y,2} \\
    X &= \left[\mbox{sin}(0.5t_x + \phi), \mbox{sin}(0.5t_x + 2*\phi\right] + \eta_x \\
    Y &= \left[\mbox{sin}(0.5t_y + \phi) + \Delta_0(T,t_y), sin(0.5*t_y + 2*\phi) + \Delta_1(T,t_y)\right]  + \eta_y \\
\end{align*}

where $\Delta_0(T)$ and $\Delta_1(T)$ are defined as follows:

\begin{equation*}
    \Delta_0(T,t) = \begin{cases}
    \frac{\mbox{min}(t-20,t_{p}-20)}{t_{p}-20} \cdot T &\text{if $U_Z=0$ or $U_Z=2$} \\
    0 &\text{if $Z=1$} \\
    \end{cases}
\end{equation*}
\begin{equation*}
    \Delta_1(T,t) = \begin{cases}
    \frac{\mbox{min}(t-20,t_{p}-20)}{t_{p}-20} \cdot T &\text{if $U_Z=1$ or $U_Z=2$} \\
    0 &\text{if $Z=0$} \\
    \end{cases}
\end{equation*}

\begin{figure}[htbp!]
    \centering
    \includegraphics[width=0.7\linewidth]{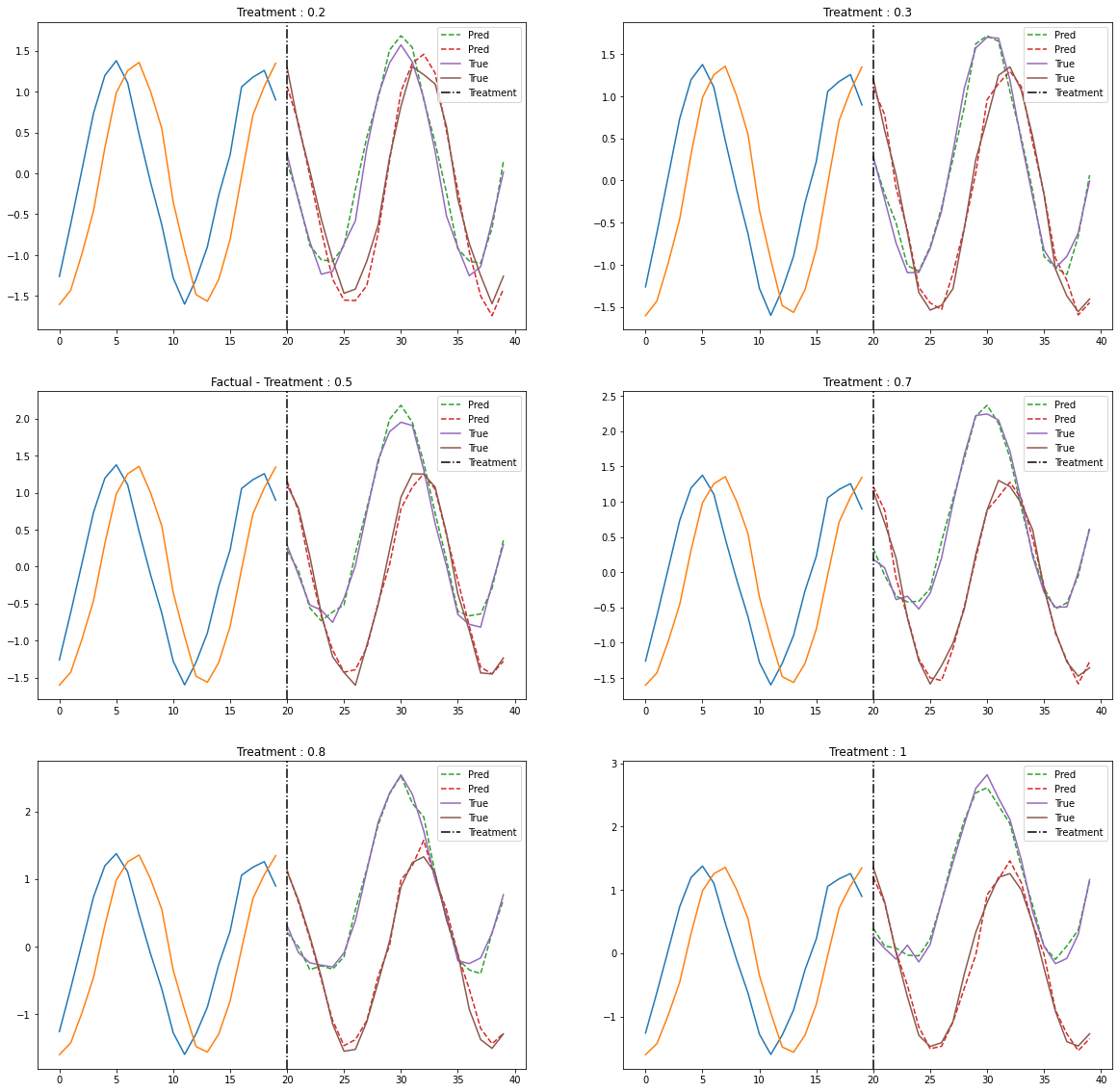}
    \caption{Counterfactuals generated by the model for different levels of treatment. The factual treatment is $T=0.5$.}
    \label{fig:cf_ts_exps}
\end{figure}

$t_{p} = 3$ characterizes the time constant of the treatment response dynamics. The latent categorical variables $U_Z$ represent hidden patient groups that drive the treatment response. In the non-additive noise-response case, we set : 

\begin{align*}
    \phi \sim \mathcal{N}(0,\sigma^2)^{40}\\
\end{align*}

such that the phase $\phi$ is time-varying and stochastic. Examples of the trajectories and reconstructed counterfactuals by \method~ are provided in Figure~\ref{fig:cf_ts_exps}.

\subsection{Cardiovascular Dataset}

We use an ODE model of the cardiovascular system as proposed in \cite{zenker2007inverse,brouwer2022predicting}. Fluid intake is commonly used for treating severe hypotension. However, the response of patient to fluids intake is difficult to assess beforehand. In particular, it depends on the patients cardiac contractility factor and the blood pressure at time of injection. If blood pressure is commonly and easily measured in standard clinical practice, assessing the cardiac contractility level of a patient requires imaging techinques such as echocardiography to measure the stroke volume. Yet, the injection of significant volume of fluids in an irreponsive patients cardiovascular system can lead to severe damage. This lead some clinicians to advocate for fluid challenges, or limited amount of fluid injection to test the responsiveness. This technique is still contested in the medical community and legs raising challenge, much less damaging but also less effective at assessing a patients response has been encouraged. This lack of availability of clear guidelines for fluids intake makes it a perfect case study for counterfactual prediction. Indeed, we'll try to address the question of if a clinician should administer fluids to particular patient based on his clinical history and therefore help informing clinical practice. The system of ODE used to generate the data is the following :

\begin{align*}
\frac{d S V(t)}{d t} &=I_{\text {external }(t)} \\
\frac{d P_{a}(t)}{d t} &=\frac{1}{C_{a}}\left(\frac{P_{a}(t)-P_{v}(t)}{R_{T P R}(S)}-S V \cdot f_{H R}(S)\right) \\
\frac{d P_{v}(t)}{d t} &=\frac{1}{C_{v}}\left(-C_{a} \frac{d P_{a}(t)}{d t}+I_{\text {external }(t)}\right) \\
\frac{d S(t)}{d t} &=\frac{1}{\tau_{\text {Baro }}}\left(1-\frac{1}{1+e^{-k_{\text {width }}\left(P_{a}(t)-P_{a_{\text {set }}}\right)}}-S\right)
\end{align*}

where

\begin{align*}
R_{T P R}(S(t)) &=S(t)\left(R_{T P R_{M a x}}-R_{T P R_{M i n}}\right) +R_{T P R_{M i n}}+R_{T P R_{M o d}} \\
f_{H R}(S(t)) &=S(t)\left(f_{H R_{M a x}}-f_{H R_{M i n}}\right)+f_{H R_{M i n}} .
\end{align*}

In the above dynamical system, $P_a, P_v, S$ and $SV$ stand for arterial blood pressure, venous blood pressure, autonomic baroreflex tone and cardiac stroke volume respectively. $I_{\text {external }(t)}$ is the amount of fluids given the patient over time and corresponds to the exogeneous input $u_T(t)$ in our model. In the data generation, we model it as 

\begin{align*}
    I_{\text {external }}(t) &= (1 + 2U_Z) \cdot T \cdot 5 \cdot f(P_a(t=0)) \cdot e^{-(\frac{t-t_{\text{treat}}-5}{5})^2}\\
    I_{\text {external }}(t) &= 0 \quad \forall t\leq t_treat\\
\end{align*}

where the treatment assignment $T$ is generated as $T\sim \mathcal{U}(0.6,1)$ and the hidden group assignment $U_Z \sim \mathcal{B}\mbox{er}(0.5)$ and $t_{\text{treat}}$ is the time of treatment. The function $f$ introduces confounding in the treatment assignment by setting :

\begin{align*}
    f(P_a(t=0)) &= g(0.5 + (P_a(t=0)-0.75)/0.1)
    g(x) &= 0.02 \cdot \left( \mbox(cos)(5x-0.2)\cdot (5-x)^2\right)^2\\ 
\end{align*}

We simulate the above system of ODEs for $t_{\text{span}}=40$ seconds and sample an observation every $\Delta_t = 1$ second. The treatment assignment time is set to $t_{\text{treat}}=20$.

\begin{figure}
    \centering
    \includegraphics[width=0.8\linewidth]{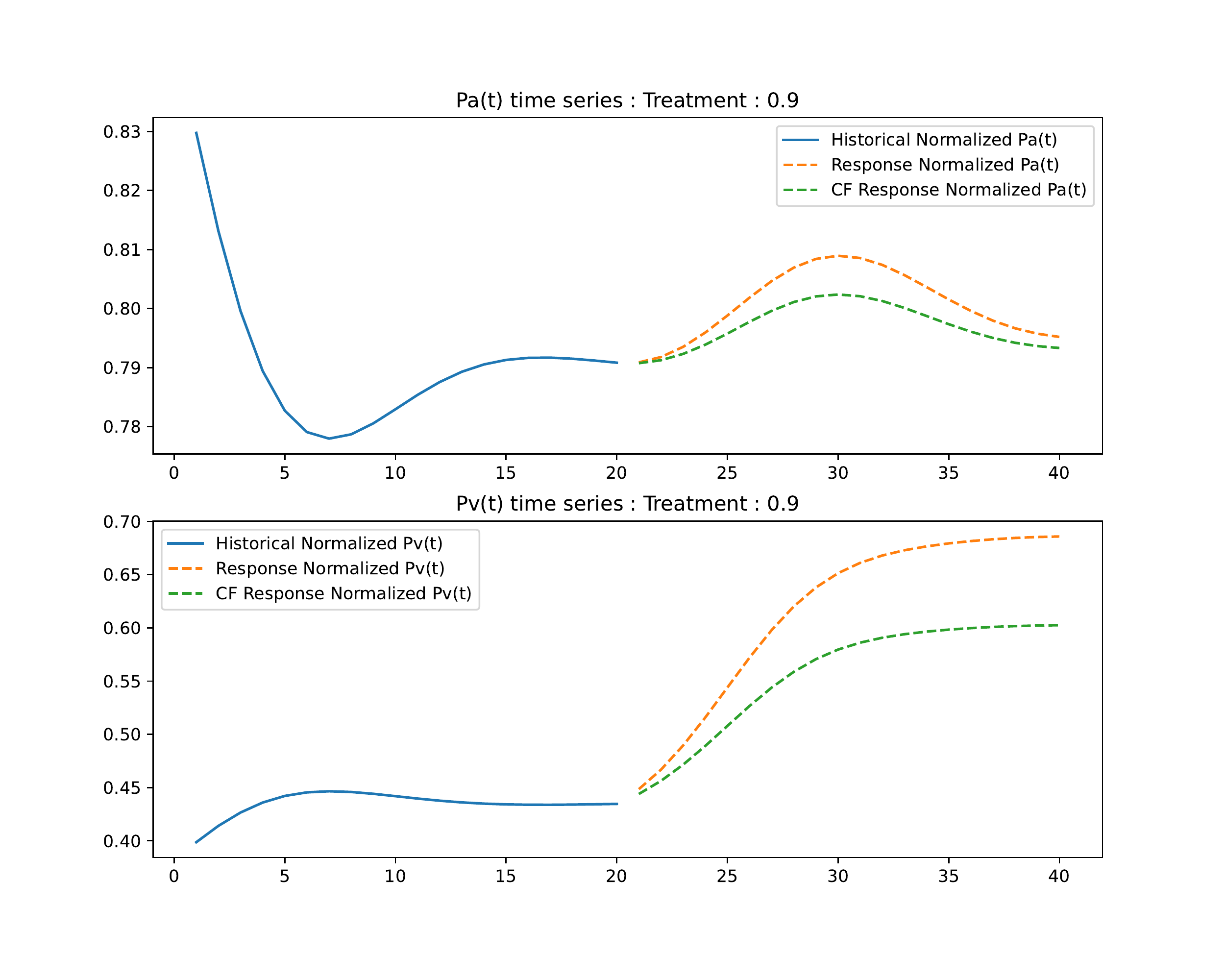}
    \caption{Example of time series in the cardio-vascular data set.}
    \label{fig:cv_example}
\end{figure}

We then set $X$ as the first two dimensions of the dynamical system ($(P_a,P_v$) for $t< t_{\text{treat}}$ and $Y$ as ($(P_a,P_v$) for $t\geq t_{\text{treat}}$. An example of a generated time series is given in Figure \ref{fig:cv_example}.

\subsubsection{Noise Responses}

\paragraph{Additive noise}

In the additive noise case, we add $U_{\eta} \sim \mathcal{N}(0,\sigma^2)^t_{\text{span}}$ on both $X$ and $Y$

\paragraph{Non-additive noise}

In the non additive case, we modify the treatment assignment by setting:

\begin{align*}
    I_{\text {external }}(t) &= (1 + 2U_Z + U_\eta(t)) \cdot T \cdot 5 \cdot f(P_a(t=0)) \cdot e^{-(\frac{t-t_{\text{treat}}-5}{5})^2}\\
\end{align*}

Where $U_{eta}(t)$ is a Gaussian process with zero-mean and diagonal covariance function with variance $\sigma^2$.

\section{Experiments Details}
\label{app:experiments}

\subsection{Computational Resources}

We ran our experiments on cluster containing two types of GPUs : NVIDIA Titan Xp and NVIDIA Quadro GV100. Our experiments resulted in 7 GPU-days of computation.

\subsubsection{Hyper-parameters}

In Table \ref{tab:hyper-params}, we report the hyper-parameters used in our experiments.

\begin{table}[htbp]
    \centering
    \begin{adjustbox}{width=\textwidth}
    \begin{tabular}{cccccccc}
    \toprule
    Hyper-parameter & Description &
    \multicolumn{2}{c}{Image} & \multicolumn{2}{c}{Harmonic Oscillator} & \multicolumn{2}{c}{Cardiovascular}  \\
    \midrule
    & & additive & non-additive & additive & non-additive & additive & non-additive \\
    \midrule
     $lr$  & learning rate & $0.001$ & $0.001$ &  $0.001$ &  $0.001$ & $0.001$ & $0.001$ \\
    $\Delta$  & update period & $10$ & $10$ & $20$ & $20$ & $20$ & $20$  \\
      $\sigma$  & noise variance & $0.01$  & $0.05$ & $0.05$ & $0.05$ & $0.01$ & $0.01$  \\  
    $\mbox{epochs}_0$  & number of epochs initialization & $50$  & $50$ & $500$ & $500$ & $500$  & $500$ \\  
    $\mbox{epochs}_1$  & number of epochs fine-tune & $50$  & $50$ & $500$ & $500$ & $500$ & $500$ \\  
    $K_0$  & number of classes & $50$  & $6$ & $6$ & $3$ & $3$ & $2$ \\  
    $\mbox{bs}$  & batch size & $128$  & $128$ & $128$ & $128$ & $128$ & $128$ \\  
    $N_{\mbox{train}}$  & Number of train samples & $42,000$  & $42,000$ & $128$ & $128$ & $500$ & $500$ \\  
    $N_{\mbox{val}}$  & Number of train samples & $9,000$  & $9,000$ & $128$ & $128$ & $250$ & $250$ \\  
    $N_{\mbox{test}}$  & Number of train samples & $1,000$  & $1,000$ & $1,000$ & $128$ & $1,000$ & $1,000$ \\  
      \bottomrule
    \end{tabular}
    \end{adjustbox}
    \caption{Hyper-parameters used for training \method~on the different datasets.}
    \label{tab:hyper-params}
\end{table}

\subsection{Number of groups}
\label{app:number_groups}

\begin{figure}[htbp]
    \centering
    \begin{subfigure}[t]{0.45\textwidth}
    \centering
    \includegraphics[width=1.1\linewidth]{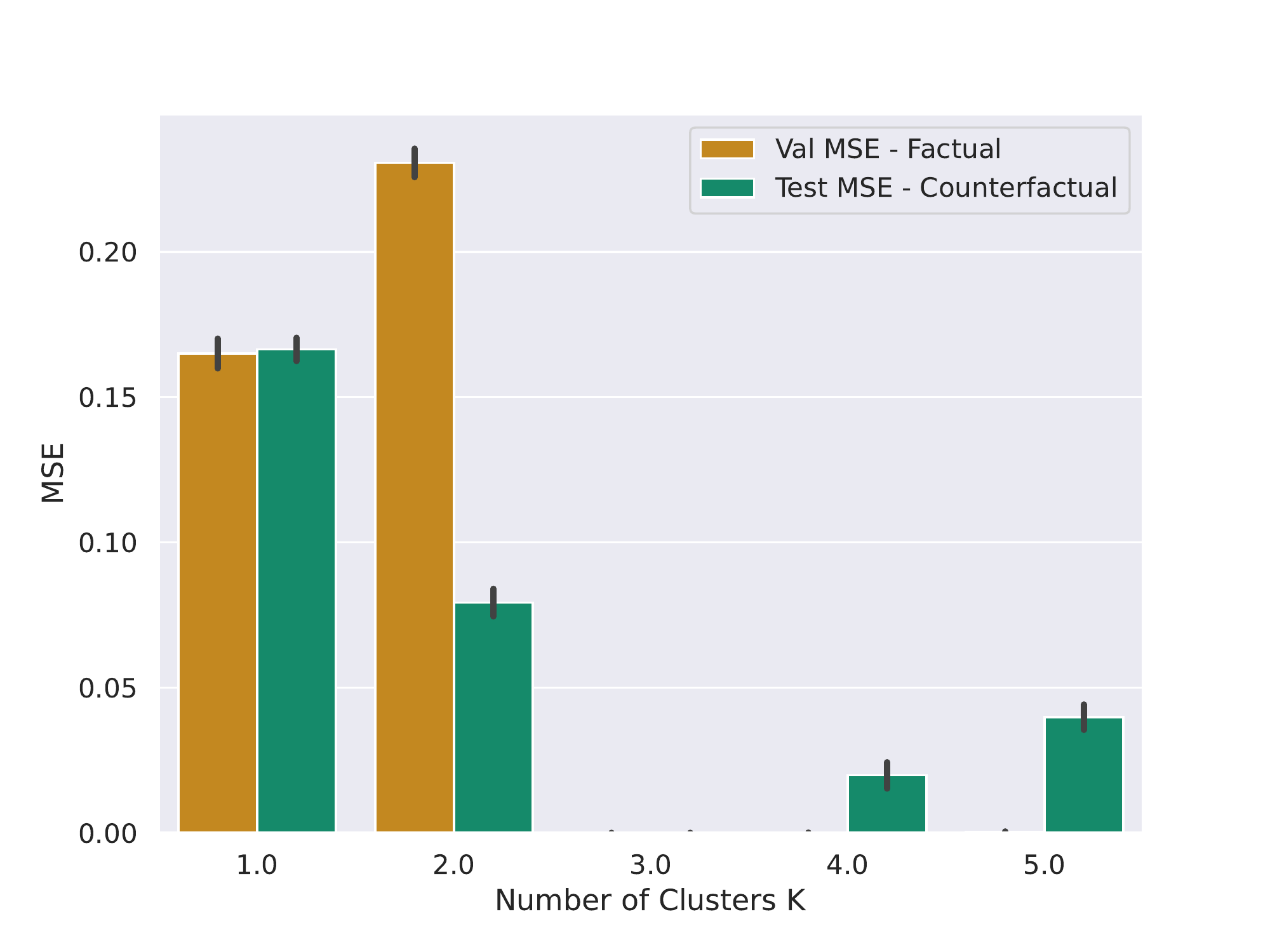}
    \caption{Reconstruction MSE in function of the number of groups in the harmonic oscillator dataset (additive case). The true number of groups is  3.}
    \label{fig:group_SimpleTraj_additive}
    \end{subfigure}
    \hfill
    \begin{subfigure}[t]{0.5\textwidth}
    \centering
    \includegraphics[width=\linewidth]{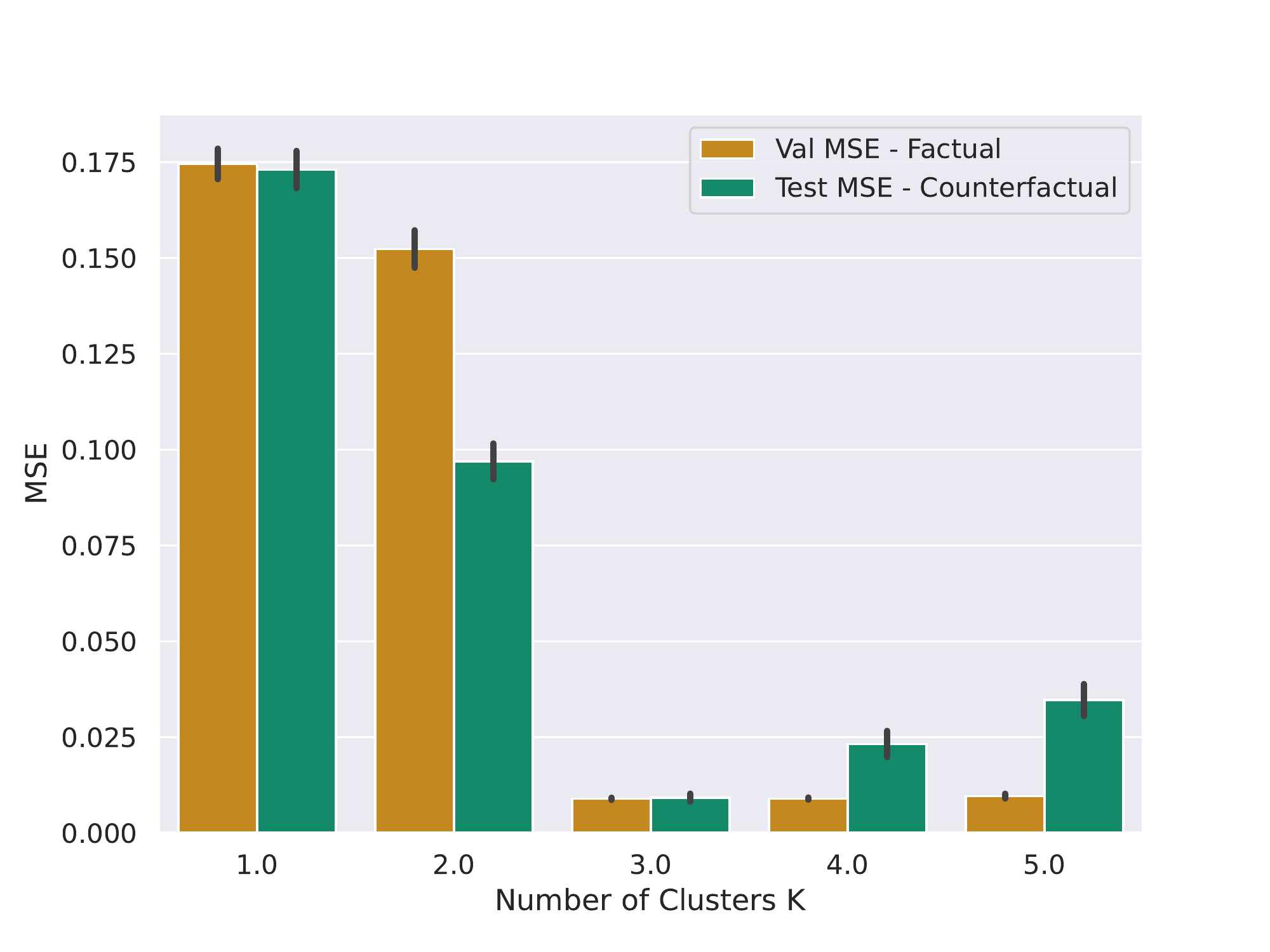}
    \caption{Reconstruction MSE in function of the number of groups in the harmonic oscillator dataset (non-additive case). The true number of groups is  3.}
    \label{fig:group_SimpleTraj_non_additive}
\end{subfigure}
\hfill
\caption{Analysis of the impact of the number of clusters $K$ on the validation and test performance for the Harmonic Oscillator dataset.}
\end{figure}

\begin{figure}[htbp]
    \centering
    \begin{subfigure}[t]{0.45\textwidth}
    \centering
    \includegraphics[width=1.1\linewidth]{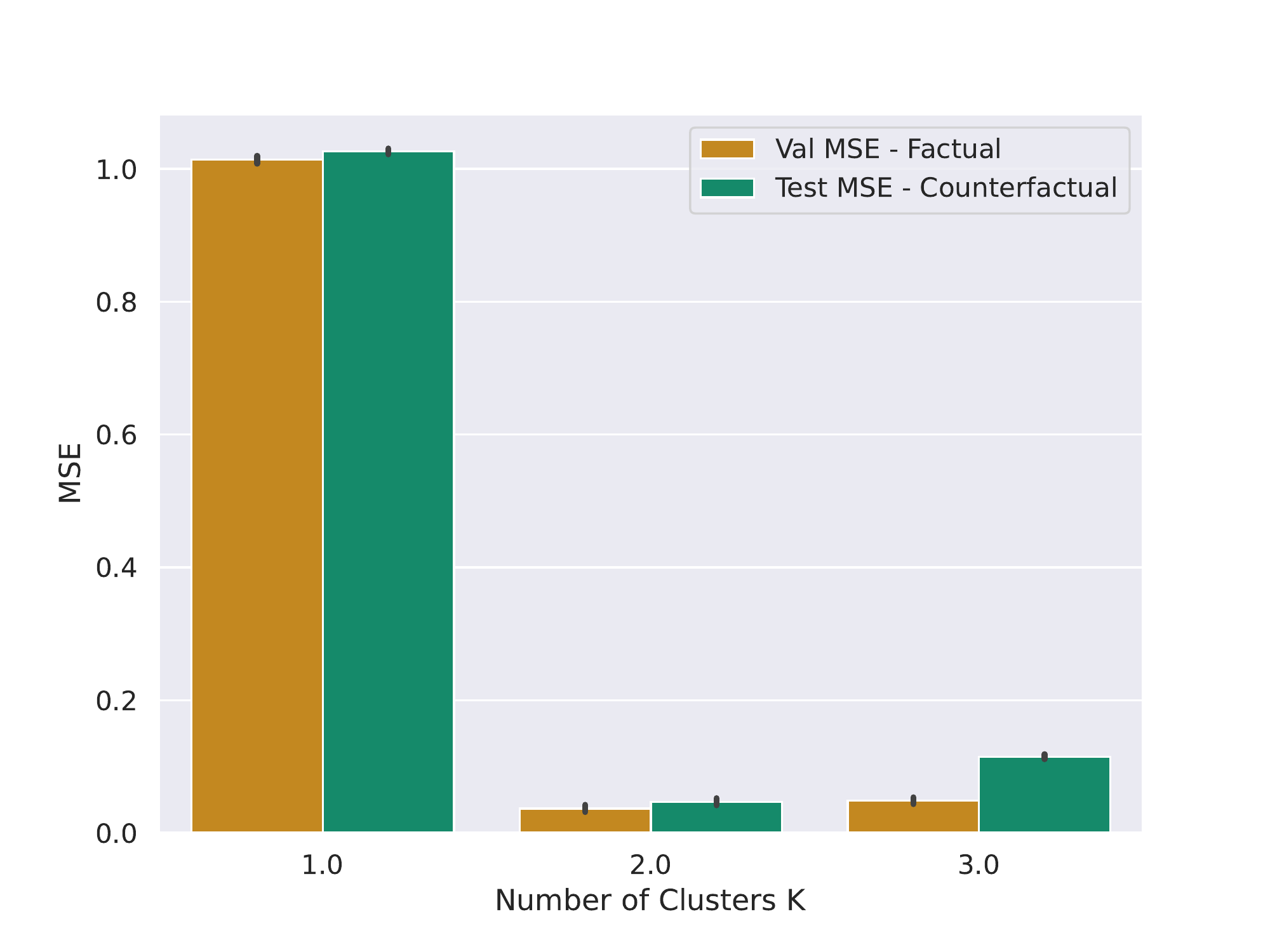}
    \caption{Reconstruction MSE in function of the number of groups in the cardiovascular dataset (additive case). The true number of groups is 2.}
    \label{fig:group_CV_additive}
    \end{subfigure}
    \hfill
    \begin{subfigure}[t]{0.5\textwidth}
    \centering
    \includegraphics[width=\linewidth]{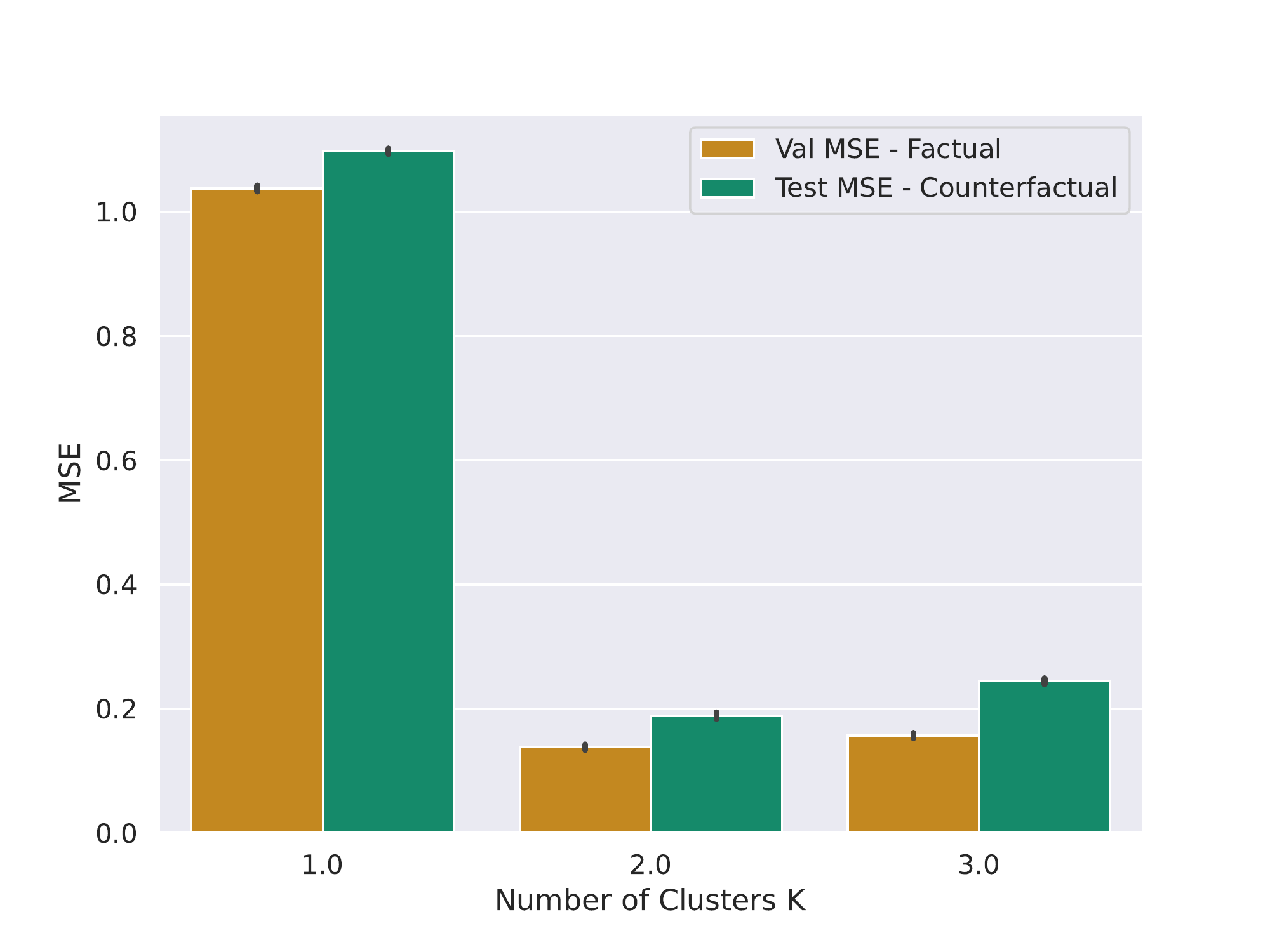}
    \caption{Reconstruction MSE in function of the number of groups in the cardiovascular dataset (non-additive case). The true number of groups is 2.}
    \label{fig:group_CV_non_additive}
\end{subfigure}
\hfill
\caption{Analysis of the impact of the number of clusters $K$ on the validation and test performance for the Cardiovascular dataset.}
\end{figure}

\begin{figure}[htbp]
    \centering
    \begin{subfigure}[t]{0.45\textwidth}
    \centering
    \includegraphics[width=1.1\linewidth]{CF_figs/numcenters_cf.pdf}
    \caption{Reconstruction MSE in function of the number of groups in the colored MNIST dataset (additive case). The true number of groups is 6.}
    \label{fig:group_MNIST_additive}
    \end{subfigure}
    \hfill
    \begin{subfigure}[t]{0.5\textwidth}
    \centering
    \includegraphics[width=\linewidth]{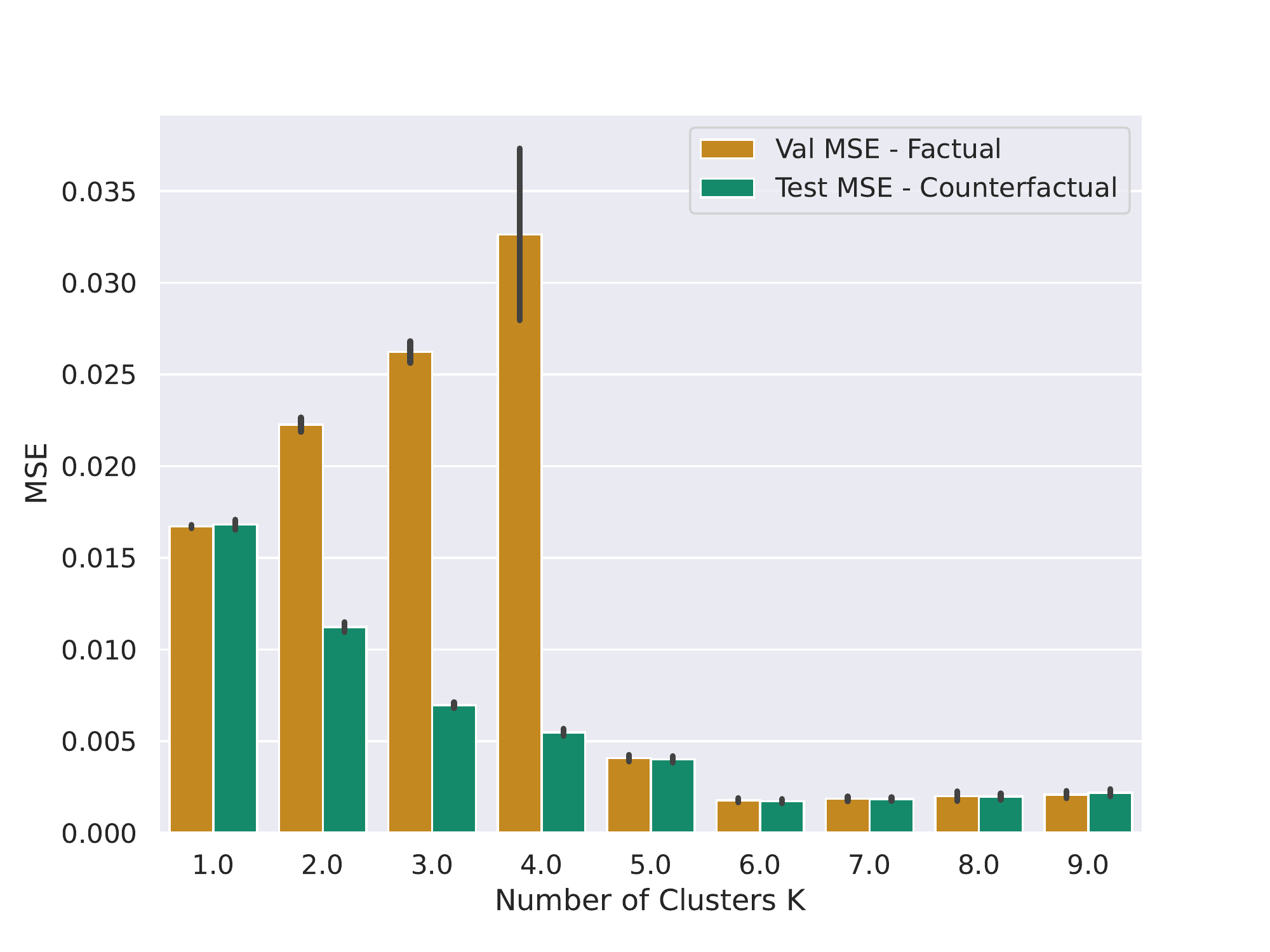}
    \caption{Reconstruction MSE in function of the number of groups in the colored MNIST dataset (non-additive case). The true number of groups is 6.}
    \label{fig:group_MNIST_non_additive}
\end{subfigure}
\hfill
\caption{Analysis of the impact of the number of clusters $K$ on the validation and test performance for the colored MNIST dataset.}
\end{figure}

As discussed in section \ref{sec:robustness}, the number of groups is an important hyper-parameter in our model. In Table \ref{tab:groups},we report the validation reconstruction loss and the corresponding test counterfactual reconstruction MSE for all datasets and different values of $K$. We observe that we obtain a lower validation error for the true value of $K$ in all datasets. In Figures \ref{fig:group_SimpleTraj_additive} and \ref{fig:group_SimpleTraj_additive}, we visualize the validation and test performance graphically for the Harmonic Oscillator dataset. We do the same on Figures \ref{fig:group_CV_additive} and \ref{fig:group_CV_non_additive} for the Cardiovascular dataset and on Figures \ref{fig:group_MNIST_additive} and \ref{fig:group_MNIST_non_additive} for the colored MNIST dataset.

\begin{table}[htbp]
    \centering
    \begin{tabular}{llccc}
\toprule
  Data & Noise & Number of Centers & Validation MSE & Test MSE  \\
\midrule
MNIST & Additive & 1 &  $0.0171\pm0.00012 $ & $0.01657\pm0.00038 $ \\

MNIST & Additive & 2 & $0.02267\pm0.00056 $ & $0.01094\pm0.00025 $\\

MNIST & Additive & 3 & $0.02746\pm0.00081 $ & $0.00677\pm9e-05 $ \\

MNIST & Additive &  4 &  $0.02552\pm0.0022 $ & $0.00529\pm1e-04 $\\

MNIST & Additive &  5 & $0.02556\pm0.01707 $ & $0.00377\pm0.00016 $\\

MNIST & Additive &  6 &  $\mathbf{0.00271\pm0.0003} $ & $0.00145\pm6e-05 $ \\

MNIST & Additive &  7 & $0.00272\pm0.00028 $ & $0.0016\pm7e-05 $\\

MNIST & Additive &  8 & $0.00278\pm0.00027 $ & $0.00167\pm7e-05 $ \\

MNIST & Additive &  9 & $0.00289\pm0.00031 $ & $0.00179\pm6e-05 $\\

\midrule

MNIST & Non-Additive &  1 &   $ 0.01673\pm9e-05 $ & $ 0.01673\pm0.00026 $ \\
MNIST & Non-Additive &  2 &  $ 0.02228\pm0.0004 $ & $ 0.01125\pm0.00026 $\\
MNIST & Non-Additive &  3 & $ 0.02632\pm0.00063 $ & $ 0.00693\pm0.00017 $\\
MNIST & Non-Additive &  4 & $ 0.03182\pm0.00653 $ & $ 0.00552\pm0.00017 $\\
MNIST & Non-Additive &  5 &  $ 0.0041\pm0.00018 $ & $ 0.004\pm0.00018 $\\
MNIST & Non-Additive &  6 & $\mathbf{ 0.00179\pm0.00014} $ & $ 0.00171\pm0.00012 $\\
MNIST & Non-Additive &  7 & $ 0.00186\pm0.00014 $ & $ 0.00185\pm0.00011 $\\
MNIST & Non-Additive &  8 & $ 0.00198\pm0.00022 $ & $ 0.00198\pm0.00018 $\\
MNIST & Non-Additive &  9 & $ 0.00212\pm0.00023 $ & $ 0.00217\pm0.00022 $\\

\midrule

Harmonic Oscillator & Additive & 1 &  $ 0.1653\pm0.01086 $ & $ 0.16584\pm0.00443 $\\
Harmonic Oscillator & Additive &  2 & $ 0.23024\pm0.04305 $ & $ 0.08058\pm0.01735 $ \\
Harmonic Oscillator & Additive &  3 &     $ \mathbf{9e-05\pm3e-05} $ & $0.01332 \pm 0.00123$ \\
Harmonic Oscillator & Additive &  4 &   $ 0.00013\pm1e-04 $ & $ 0.0199\pm0.00604 $\\
Harmonic Oscillator & Additive & 5 & $ 0.00035\pm0.00024 $ & $ 0.04017\pm0.00697 $\\

\midrule

Harmonic Oscillator & Non-additive &  1 & $ 0.17451\pm0.01053 $ & $ 0.17459\pm0.00704 $ \\
Harmonic Oscillator & Non-additive &  2 & $ 0.15164\pm0.10242 $ & $ 0.09669\pm0.01678 $ \\
Harmonic Oscillator & Non-additive &  3 & $ \mathbf{0.00891\pm0.00035} $ & $ 0.00956\pm0.00096 $ \\
Harmonic Oscillator & Non-additive &  4 & $ 0.00905\pm0.00032 $ & $ 0.02279\pm0.00327 $  \\
Harmonic Oscillator & Non-additive &  5 & $ 0.00984\pm0.00073 $ & $ 0.03464\pm0.01158 $ \\

\midrule

CardioVascular & Additive & 1 & $ 1.01453\pm0.06228 $ & $ 1.02569\pm0.04794 $ \\
CardioVascular & Additive &  2 & $ 0.03677\pm0.01694 $ & $ 0.07701 \pm 0.05001 $\\
CardioVascular & Additive &  3 & $ 0.04906\pm0.02931 $ & $ 0.11363\pm0.06759 $ \\

\midrule

CardioVascular & Non-additive &  1 & $ 1.03865\pm0.12068 $ & $ 1.09768\pm0.05459 $\\
CardioVascular & Non-additive &   2 &  $\mathbf{ 0.1372\pm0.05497 }$ & $ 0.18797\pm0.11421 $ \\
CardioVascular & Non-additive &   3 & $ 0.15747\pm0.10191 $ & $ 0.2434\pm0.13175 $\\

\bottomrule
\end{tabular}
    \caption{Reconstruction validation MSE and counterfactual test MSE for all datasets and different number of groups ($K$). Lowest validation MSE are bolded for each dataset.}
    \label{tab:groups}
\end{table}

\subsection{Strength of Correlation Experiment}

Our approach holds even when the latent background variable $U_Z$ and the covariates $X$ are correlated. When this is the case, information from $X$ can be used to infer the value of $U_Z$. In Figure \ref{fig:correlation}, we show the impact of the correlation strength $\rho$ between $X$ and $U_Z$. 

We incorporate correlation between $X$ and $U_Z$ as follows. The base probability $p_0^k$ of each $U_Z$ is uniform. That is,

\begin{align*}
    P^0(U_Z = k) = p_0^k = \frac{1}{K}
\end{align*}

We modify this probabilty based on the label of the image (the digit label $y_{\mbox{label}} \in [9]$. For a correlation factor $\rho \in [0,1]$, we write

\begin{align*}
    p^k = 
    \begin{cases}
    \frac{1-((1-p_0)\rho+p_0)}{K-1} \quad \text{if } k \neq y_{label} \mbox{mod} (K) \\
     ((1-p_0)\rho+p_0) \quad \text{if } k = y_{label} \mbox{mod} (K) \\
    \end{cases}
\end{align*}

The above equation suggest that some images with specific label classes will have higher probability of sampling a latent class $U_Z = u_Z$ such that $y_{\mbox{label}}\mbox{mod}(K)=u_Z$.

In the case when $\rho=0$, $p^k=p_0^k=\frac{1}{K}$, corresponding to a uniform probability of latent class for each class label. When $\rho=1$, $p^{y_{\mbox{label}\mbox{mod}(K)}}=1$ and the other probabilities are set to $0$, corresponding to a deterministic assignment of $U_Z$ conditioned on $y_{\mbox{label}}$.

\subsection{Scalability}

\method scales linearly with the number of groups $K$, though it can be easily parallelized. Indeed, each sub-model $m_i$ can be evaluated in parallel. Regarding scaling with respect to the dimensions of the outcomes and treatments, we can use arbitrary neural networks architectures to process them, for which the complexity can vary. In our experiments, we used CNNs for images, which scale quadratically with the size of the images and TemporalCNNs for time series which scale linearly with the length of the time series as well as number dimensions.

\subsection{Clustering Algorithm}
\label{app:clustering}

Our approach relies on an \emph{initial} clustering step as described in Section \ref{sec:cfqp}. In our main experiments, we use a k-means algorithm for this step. However, different clustering strategies can be considered. In Table \ref{tab:gmm}, we report the performance metrics of a variant of our approach using Gaussian Mixture Models (GMM). For computational reasons, the GMM is fit by using a random sub-sample of the training data (1000 samples). We observe only minor performance deviations between k-means and GMM versions of \method. 

\begin{table}[ht!]
\centering
\caption{Test MSE of the counterfactual reconstructions for the different datasets.}
\begin{adjustbox}{width=\textwidth}
\begin{tabular}{lccc|ccc}
\toprule[1.5pt]
  & \multicolumn{3}{c}{Additive Noise} & \multicolumn{3}{c}{Non-Additive Noise} \\
  \midrule
Model &  Harmonic Oscillator & colored-MNIST & Cardiovascular & Harmonic Oscillator & colored-MNIST & Cardiovascular\\
\midrule[1.5pt]
Deep-ITE~\cite{johansson2016learning} & $0.187 \pm 0.006$ & $0.017 \pm 0.001$ & $ 1.084\pm0.087 $ & $ 0.174\pm0.004 $ & $ 0.017\pm0.001 $ & $ 1.14\pm0.121 $\\
\midrule
SC~\cite{abadie2021using} & $0.177 \pm 0.131$ & $ 0.020\pm0.001 $ & $ 1.610\pm0.141 $ & $ 0.167\pm0.004 $ & $ 0.020\pm0.001 $ & $ 1.628\pm0.144 $\\
\midrule
Deep-SCM~\cite{pawlowski2020deep} & $0.124 \pm 0.005$ & $0.011 \pm 0.001$  &  $0.405 \pm 0.042$ & $0.123 \pm 0.005$ & $0.011 \pm 0.001$ & $0.424 \pm 0.042$\\
\midrule
Diff-SCM~\cite{sanchez2021diffusion} & $0.082 \pm 0.023$ & $0.008 \pm 0.004$ & $ 0.206\pm0.036 $ & $ 0.106\pm0.038 $ & $ 0.009\pm0.002 $ & $ 0.311\pm0.073 $ \\
\midrule
\method~(ours) & $ 0.013 \pm 0.001$ & $\mathbf{0.001 \pm 0.001}$ & $0.077 \pm 0.050$ & $ \mathbf{0.009\pm0.001} $ & $ \mathbf{0.002\pm0.001} $ & $ \mathbf{0.188\pm0.114} $ \\
\midrule
\method -GMM~(ours) & $\mathbf{0.001 \pm 0.001}$ & $0.002 \pm 0.001$ & $ \mathbf{0.067\pm0.042}$  & $\mathbf{0.009\pm0.001} $ & $\mathbf{0.002 \pm 0.001}$ & $ 0.192\pm0.0834 $ \\
\bottomrule
\end{tabular}
\end{adjustbox}
\label{tab:gmm}
\end{table}

\section{Proof of Result~\ref{result}}
\label{app:proofs}

Our proof is structured as follows. We first show identifiability of non-parameteric mixtures at a single point ($(X=x,T=t)$. We then show identifiability of the permutation function over $(\mathcal{X}\times \mathcal{T})$. Finally, based on these results, we derive the bound of Result~\ref{result}.

\subsection{Identifiability of Non-Parametric Mixture Models.}
\label{app:nonparametric}

To show the identifiability in the non-additive case, we re-use notation and start from the outline derived in \citet{aragam2020identifiability}.

\subsubsection{Identifiability Result}

We let $(X,d)$ be a metric space and ($\mathcal{P}(X),\rho$) the space of regular Borel probability measures on $X$ with finite $r$th moments metrized by a metric $\rho$. Let's further define $\mathcal{P}^2(X) = \mathcal{P}(\mathcal{P}(X))$, the space of mixing measures over $\mathcal{P}(X)$. 
We represent a mixture distribution over $X$ as a new probability measure $m(\cdot,\Lambda) \in \mathcal{P}(X)$ such that $m(A;\Lambda) = \int \gamma(A) d\Lambda(\gamma)$ for any $A \subset X$.

Given a Borel set $\mathfrak{L} \subset \mathcal{P}^2(X)$, we define $\mathcal{M}(\mathfrak{L}) := \{ m(\Lambda) : \Lambda \in \mathfrak{L})$ as a family of mixture distributions over $X$. Because we are interested in finite mixtures, we use $\mathcal{P}_k^2 := \{ \Lambda \in \mathcal{P}^2(X) : \lVert \mbox{supp}(\Lambda)\rVert \leq k\}$ to denote the set of mixing measures with $k$ components. Finite mixtures $\Gamma \in \mathcal{M}(\mathcal{P}_k^2(X))$ can thus also be written in a more intuitive form, \emph{i.e.}  $\Gamma = \sum_{k=1}^K \omega_k \gamma_k$ where $\gamma_k \in \mathcal{P}(X)$ are the \emph{mixture components} and $\omega_k$ are mixing weights, as in Equation~\ref{eq:mixture} in the main text.

Note that we did not make any assumption regarding the distribution of the mixture components ($\gamma_k \in \mathcal{P}(X)$). Now, from \cite{aragam2020identifiability}, we have the following result.

\begin{realtheorem}
If $\mathfrak{L}$ is a $\mathfrak{Q}_{L}$-clusterable family, then there exists a function $h$ : $\mathcal{M}(\mathfrak{L}) \rightarrow \mathfrak{L}$ such that $h(m(\Lambda))=\Lambda$, where $m: \mathfrak{L} \rightarrow \mathcal{M}(\mathfrak{L})$ is the canonical embedding. In particular, $m$ is a bijection and the mixture model $\mathcal{M}(\mathfrak{L})$ is identifiable.
\end{realtheorem}

We see that identifiability of the mixture therefore hinges on the notion of \emph{clusterability} of the individual distributions that compose the mixture. We now make the notion of clusterability more precise.

\subsubsection{Clusterability}

\paragraph{Projections}

Let's first introduce $\{\mathfrak{Q}_L\}_{L=1}^{\infty}$ as an indexed collection of families of mixing measures that satisfy the following requirements:
\begin{itemize}
    \item $\mathfrak{Q}_L \subset \mathcal{P}^2_L(X)$ for each $L$;
    \item $\{ \mathfrak{Q}_L \}$ is a filtration ($\mathfrak{Q}_L \subset \mathfrak{Q}_{L+1}$);
    \item The collection of mixture distributions $\mathcal{M}(\mathfrak{Q}_L)$ is identifiable for each $L$.
\end{itemize}

Examples of such a collection is the filtration of Gaussian mixtures with $L$ components.

We then proceed by defining the $\rho$-projection of a mixture distribution $\Gamma = m(\Lambda)$ on a family of mixture distribution $\mathfrak{Q}_L$ by

\begin{align*}
    T_L{\Gamma} = \left\{ Q \in \mathfrak{Q}_L: \rho(Q,\Gamma) \leq \rho(P,\Gamma) \forall P \in \mathcal{\mathfrak{Q}_L}\right\}
\end{align*}

where $\rho$ is a metric on $\mathcal{P}(X)$ (\emph{e.g.} the Hellinger distance). This projection intuitively maps a mixture distribution $\Gamma$ to the closest element of the family $\mathfrak{Q}_L$ it is projected upon. We futher define the map $M_L : \mathcal{\mathfrak{Q}_L} \rightarrow \mathfrak{Q}_L$ that a projected distribution to its mixing measure:

\begin{align*}
M_{L}\Gamma = M_L(T_L\Gamma)= \left\{\Omega \in \mathfrak{Q}_{L}: m(\Omega) \in T_{L} \Gamma\right\}
\end{align*}

\paragraph{Assignment Functions}

Because it is an element of $\mathfrak{Q}_L$, the projection $T_L\Gamma = Q^*$ is a mixture distribution and can thus be written as $T_L\Gamma = \sum_{l=1}^L \omega^*_l \gamma^*_l$. When projecting on mixture distributions with more mixture components than in the original distribution, \emph{i.e.} $L \geq K$, one can then assign mixture components in $Q^*$ to mixture components in $\Gamma$. We define the set of assignment maps $\alpha : [L] \rightarrow [K]$ as $\mathbb{A}_{L \rightarrow K}$. Given such an assignment map, we define $\psi_k(\alpha) := \sum_{l \in \alpha^-1(k)} \omega_l$ for the weights induced by the assignment map and $Q_k(\alpha) := \frac{1}{\psi_k(\alpha)} \sum_{l \in \alpha^{-1}(k)} \omega^*_l \gamma^*_l$ for the induced mixing components.

\paragraph{Regularity of a mixing measure}
We are now ready for introducing the definition of regularity of a mixing measure.

\begin{realdefinition}[Regularity]
Suppose $\Lambda \in \mathcal{P}_{K}^{2}(X)$ and $\Gamma=m(\Lambda) \in \mathcal{M}_{K}(\mathcal{P}^2(X))$. The mixing measure $\Lambda$ is called $\mathfrak{Q}_{L}$-regular if:
\begin{itemize}
    \item(a) The $\rho$-projection $Q^{*}=T_{L} \Gamma$ exists and is unique for each $L$ and $\lim _{L \rightarrow \infty} T_{L} \Gamma=$ $\Gamma$;
    \item(b) There exists a sequence of assignment functions $\alpha=\alpha_{L} \in \mathbb{A}_{L \rightarrow K}$ such that $\lim _{L \rightarrow \infty} Q_{k}^{*}(\alpha)=\gamma_{k} \quad$ and $\quad \lim _{L \rightarrow \infty} \psi_{k}^{*}(\alpha)=\omega_{k} \quad \forall k=1, \ldots, K$
\end{itemize}
\end{realdefinition} 

Any assignment function where (b) holds is called a \emph{regular assignment}.

\paragraph{Clusterability}

We can now finally define clusterability of a family of mixing measures with respect to another regular family of mixing measures.

\begin{realdefinition}[Clusterable family]
A family of mixing measures $\mathfrak{L} \subset \mathcal{P}^{2}(X)$ is called a $\mathfrak{Q}_{L}$-clusterable family, or just a clusterable family, if
\begin{itemize}
    \item (a) $\Lambda$ is $\mathfrak{Q}_{L}$-regular for all $\Lambda \in \mathfrak{L}$;
 \item (b) For all sufficiently large $L$, there exists a function $\chi_{L}: M_{L}(\mathfrak{L}) \rightarrow \mathbb{A}_{L \rightarrow K}$ such that $\chi_{L}\left(\Omega^{*}\right)$ is a regular assignment for every $\Lambda \in \mathfrak{L}$, with $\Omega^{*} = M_L(T_L\Gamma)$.
\end{itemize}
The resulting mixture model $\mathcal{M}(\mathfrak{L})$ is called a clusterable mixture model.
\end{realdefinition}

More intuitively, a family of mixing measures will then be $\mathfrak{Q}_{L}$-clusterable if one can project every mixing measure of the family onto $\mathfrak{Q}_{L}$ and if, for each of these projections, there exists a function that outputs a regular assignment for each mixture distribution given as input. This regular assignment is such that it clusters the elements of $Q^*$ in a way that the distribution of each cluster converges to the distribution of a element of the original mixture distribution. In this case, one can \emph{identify} a family of mixing measures by projecting each mixing distribution onto a regular family (\emph{e.g.} Gaussian mixtures) and cluster it accordingly to $\chi_L$. We refer the interested reader to \cite{aragam2020identifiability} for more insight on clusterability and identifiability in non-parametric mixtures.

\subsection{Proof of result \ref{result}}

\subsubsection{Identifiability at a Fixed Point}

We start our proof by recalling the identifiability result in the case of clusterable families of mixing measures, whose proof can be found in \cite{aragam2020identifiability}.

\begin{realtheorem}
If $\mathfrak{L}$ is a $\mathfrak{Q}_{L}$-clusterable family, then there exists a function $h$ : $\mathcal{M}(\mathfrak{L}) \rightarrow \mathfrak{L}$ such that $h(m(\Lambda))=\Lambda$, where $m: \mathfrak{L} \rightarrow \mathcal{M}(\mathfrak{L})$ is the canonical embedding. In particular, $m$ is a bijection and the mixture model $\mathcal{M}(\mathfrak{L})$ is identifiable.
\label{theorem:identifiability_theorem}
\end{realtheorem}

As laid out in the main text, we assume three observed random variables $X\in \mathcal{X}$, $Y \in \mathcal{Y}$ and $T \in \mathcal{T}$. As suggested by Assumption \ref{ass:categorical} in the main text, the conditional distribution of $Y$ at each point $(X=x,T=t)$ is given by an unknown mixture distribution from a family $\mathcal{M}_K(\mathfrak{L})$ where each mixing measure $\Lambda \in \mathfrak{L}$ satisfies $\Lambda \in \mathcal{P}^2(Y)$ and $\mid \mbox{supp}(\Lambda)\mid = K$. We write the distribution of $Y$ conditioned on $X$ and $T$ as:

\begin{align}
    Y\mid X,T \sim \gamma(X,T) = \sum_{k=1}^{K} \omega_k \gamma_{k}(X,T) \quad \text{with }  \gamma(X,T) \in \mathfrak{L}
    \label{eq:app:mixture}
\end{align}

Assuming that $\mathfrak{L}$ is clusterable, we can use Theorem \ref{theorem:identifiability_theorem} to deduce that the mixture distribution $\gamma(X,T)$ is identifiable at each fixed point $(X=x,T=t)$. 

Importantly, the idenfitifiablity insured by Theorem \ref{theorem:identifiability_theorem} is up to a permutation of the mixture components. We write $\sigma_{x,t} : [K] \rightarrow [K]$ the permutation function at a specific point $(X=x,T=t)$. We define the assignment function that maps the elements of a mixture distribution $\gamma(X,T)$ to $U_Z$ as $F_{(x,t)} : \mathcal{M}(\mathfrak{L}) \rightarrow \{ [K], \mathcal{P}(Y)\}^K$. We now move to showing that the assignment function is identifiable up to a constant permutation $\bar{\sigma}$.

\subsubsection{Identifiability of the Assignment Function}

The support of random variables $X$ and $T$, $\mathcal{X} \times \mathcal{T}$ is connected. Assuming further (Assumption \ref{ass:clusterability}) that
$\mathfrak{L}$ is a clusterable family, we know that there exists a cluster function
$\chi_{L,(x,t)}: \mathcal{M_L}(\mathfrak{L}) \rightarrow \mathbb{A}_{L\rightarrow K}$ that maps components of the projected mixture onto the initial one. We can then define a combined operator that take a mixture distribution as input and returns an indexed sequence of mixture components.

Let $F_{(x,t)} : \mathcal{M}(\mathfrak{L}) \rightarrow \{ [K], \mathcal{P}(Y)\}^K$ be this operator consisting of applying a $\rho$-projection onto a regular family, cluster the result according to $\chi_{L,(x,t)}$ and return the indexed estimated distributions $Q_k^*(\alpha)$. As the per the definition of regular mixing measures, we have also have $\lim_{L \rightarrow \infty} Q_{k}^{*}(\alpha)=\gamma_{k}$. Note that the indexing of the elements of the mixing distribution is arbitrary at a single point $(X=x,T=t)$ and thus defined up to a permutation $\bar{\sigma}$.

Because $X$ and $T$ are continuous on a connected domain, we can evaluate the operator at $(X=x+\delta x,T=t+\delta_t)$. 

\begin{align*}
F_{(x,t)} &= \{ (\sigma_{x,y}(i), q_i^*(x,t)) : i = 1,...,K\} \\
F_{(x+\delta x,t + \delta t)} &= \{ (\sigma_{x+\delta x,y + \delta y}(i), q_i^*(x+\delta x,t + \delta t)) : i = 1,...,K\} \\
\end{align*}

What is more, for arbitrary small $(\delta x, \delta t)$ and $ \forall (x,t) \in (\mathcal{X}\times \mathcal{T}); \forall k,k',k'' \in [K]$, we have 

\begin{align*}
    \rho(\gamma_k(x,t),\gamma_{k'}(x,t)) > \rho(\gamma_{k''}(x,t),\gamma_{k''}(x + \delta x,t + \delta t) \\
\end{align*}

because we require the moments of each $\gamma_k$ to be continuous in $(x,t)$. It therefore implies a relation between $\sigma_{x,y}$ and $\sigma_{x+\delta x,y + \delta y}$ that satisfies:

\begin{align*}
    \sigma_{x+\delta x,y + \delta y}(i) = \mbox{argmin}_j \rho(q_i^*(x,t),q_j^*(x+\delta x,t + \delta t))
\end{align*}

The assignment function $F_{(x,t)}$ is thus identifiable up to a constant permutation $\bar{\sigma}$ that will determine all other permutations in $\mathcal{X}\times \mathcal{T}$.

\subsubsection{Counterfactual Identifiability}

To recapitulate, in the previous sections, we have shown (1) identifiability of a non-parametric mixture at a point $(X=x,T=t)$ based on the Assumption \ref{ass:clusterability} and (2) identifiability of the permutation function over $(\mathcal{X}\times \mathcal{T})$. That is, for each point in $(X=x,T,t)$, we can identify the mixture components and the mixture weights and associate it to any other point $(X=x',T=')$, through the assignment function $F_{(x,t)}$.

In particular, we can identify the means $\mu_k$ of each cluster component $\gamma_k$. We have 

\begin{align}
    \lim_{N\rightarrow \infty} \hat{\mu}_k(X,T) = \mathbb{E}_{Y\sim \gamma_k(X,T)}\left[Y\right] = \mu_k(X,T)
\end{align}

and

\begin{align}
    \lim_{N\rightarrow \infty} \hat{\omega}_k(X,T) = \omega_k(X,T).
\end{align}

Because the above estimators converge, the following posterior probability converges to the true value as well:

\begin{align}
    &P(U_Z = u_Z \mid Y=y, X=x, T=t) \\
    &= \frac{P(Y=y\mid U_Z = u_Z, X=x, T=t)P(U_Z=u_Z,\mid X=x, T=t)}{\sum_{k=1}^K P(Y=y\mid U_Z = k, X=x, T=t)P(U_Z=k,\mid X=x, T=t)}
\end{align}

for all $x \in \mathcal{X}$, $t\in \mathcal{T}$ and $y\in \mathcal{Y}$. Indeed, $P(Y=y\mid U_Z = u_Z, X=x, T=t)$ is identifiable, and $P(U_Z=u_Z,\mid X=x, T=t) = \omega_{u_Z}(x,t)$.

Let $\hat{\omega}_{u_Z}(x,y,t)$ be the estimator of $P(U_Z = u_Z \mid Y=y, X=x, T=t)$ We are now ready to define the estimator for the counterfactual distribution $\nu_{t'}(x,y,t)$.

\begin{align}
    \nu_{t'}(x,y,t) = \sum_k \hat{\omega}_k(x,y,t) \delta(y=\hat{\mu}_k(x,t')).
\end{align}

This estimator is a discrete mixture distribution with mass on the means of the different mixture components, weighted by the posterior probability of the initial sample $(x,t,y)$ belonging to a particular component.

\subsubsection{Bounds}

We can now use this estimator $\nu_{t'}(x,y,t)$ to derive the bounds of Result \ref{result}. We first write the $W_1$ distance between the estimated and true counterfactual distributions:

\begin{align}
    & W_1(\nu_{t'}(x,t,y),Y'(x,t,y))  \nonumber\\
\end{align}

with 
\begin{align}
\nu_t'(x,t,y) &= \sum_{k=1}^N P(U_Z = k \mid X = x,T = t,Y = y) \delta(\nu_{t'}=\hat{\mu_k}(X=x,T=t')) \\
Y'(x,t,y) &= \sum_{k=1}^K P(U_Z = k \mid X=x,T=t,Y=y) \\
&\quad \quad \cdot P(Y'=y' \mid X=x,T=t,Y=y, U_Z = k)
\end{align}

We thus need to compute the $W_1$ distance between two mixture distributions where each component is scaled by the same factor. By restricting the transport map to assigning each mixture component $k$ of $\nu'_t$ to the respective one in $Y'$, we have

\begin{align}
W_1(\nu_{t'}(x,t,y),Y'(x,t,y)) \\
\leq \sum_{k=1}^K P(U_Z = k \mid x,t,y) &W_1(\delta(\nu_{t'}=\hat{\mu_k}(x,T=t')), P(Y'=y' \mid x,t,y, U_Z = k)). 
\end{align}

Now, we can write the individual components of the above sum as follows

\begin{align*}
   &P(U_Z = k \mid x,t,y) W_1(\delta(\nu_{t'}=\hat{\mu_k}(x,T=t')), P(Y'=y' \mid x,t,y, U_Z = k)) \\
   &\quad \leq P(U_Z = k \mid x,t,y) \\
   &\quad \quad \cdot \iint\limits_{y',\nu_{t'}} \lVert y' - \nu_{t'}(x,t,y) \rVert_2 \delta(\nu_{t'}=\hat{\mu_k}(x,T=t')) P(Y'=y' \mid x,t,y, U_Z = k)) \,dy' \,d\nu_{t'} \\
   & \quad = P(U_Z = k \mid x,t,y) \int\limits_{y'} \lVert y' - \hat{\mu_k}(x,T=t') \rVert_2  P(Y'=y' \mid x,t,y, U_Z = k)) \,dy'  \\
   & \quad = \int\limits_{y'} \lVert y' - \hat{\mu_k}(x,T=t') \rVert_2  P(Y'=y' \mid x,t,y, U_Z = k)) \\
   &\quad \quad \cdot \frac{P(Y \mid x,t,U_Z = k) P(U_Z = k \mid x,t)}{P(Y\mid x,t)} \,dy'  \\
\end{align*}

where we bounded the $W_1$ distance by using the joint probability distribution as the transport map between $\nu_{t}'$ and $Y'$. 

Because the $W_1$ distances are positive, the inequality applies to the expectation and we can write

\begin{align*}
&\mathbb{E}_Y[W_1(\nu_{t'}(x,t,Y),Y'(x,t,Y))] \leq \\
& \sum_{k=1}^K  \iint\limits_{y',y} \lVert y' - \hat{\mu_k}(x,T=t') \rVert_2  P(Y'=y' \mid x,t,Y, U_Z = k)) \cdot \nonumber \\
&\quad \quad \frac{P(Y \mid x,t,U_Z = k) P(U_Z = k \mid x,t)}{P(Y\mid x,t)}  P(Y\mid x, t) \,dy' \,dy
\end{align*}

We finally bound the above expectation by marginalizing $Y$:

\begin{align}
&\mathbb{E}_Y[W_1(\nu_{t'}(x,t,Y),Y'(x,t,Y))] \leq \\
& \sum_{k=1}^K  \iint\limits_{y',y} \lVert y' - \hat{\mu_k}(x,T=t') \rVert_2  P(Y'=y' \mid x,t,Y, U_Z = k) \cdot \nonumber \\
&\quad \quad P(Y \mid x,t,U_Z = k) P(U_Z = k \mid x,t) \,dy' \,dy \\
&= \sum_{k=1}^K P(U_Z = k \mid x,t) \int\limits_{y'} \lVert y' - \hat{\mu_k}(x,T=t') \rVert_2  P(Y'=y' \mid x,t, U_Z = k) \,dy' \\
&= \sum_{k=1}^K P(U_Z = k \mid x,t) \int\limits_{y'} \lVert y' - \hat{\mu_k}(x,T=t') \rVert_2  \gamma_k(y' \mid x,t, U_Z = u_Z) \,dy' \\
&= \sum_{k=1}^K P(U_Z = k \mid x,t) W_1^k(\hat{\mu}_k,\gamma_k) 
 \label{eq:combination}\\
&\leq \mbox{max}_k   W_1^k(\hat{\mu}_k,\gamma_k) \label{eq:convex}\\
&= \delta \nonumber
\end{align}

Where \ref{eq:convex} holds because it \ref{eq:combination} is a convex combination $W_1^k(\hat{\mu}_k,\gamma_k)$. This ends the proof of Result \ref{result}.

\subsubsection{The additive noise case}

Result \ref{result} mentions the bounds reduces to zero $\delta=0$ when the noise is additive. From identifiability at of a mixture at a single point, we had:

\begin{align}
    \lim_{N\rightarrow \infty} \hat{\mu_k}(x,t) = \mathbb{E}_{Y\sim \gamma_k(x,t)}\left[Y\right] = \mu_k(x,t), \label{eq:mu_convergence}
\end{align}


\begin{align}
    \lim_{N\rightarrow \infty} \hat{\omega}_k(x,t,y) = P(U_Z = u_Z \mid x,t,y). \label{eq:omega_convergence}
\end{align}

We also have convergence of the covariance matrix estimator

\begin{align}
    \lim_{N\rightarrow \infty} \hat{\Sigma_k}(x,t) = \mathbb{V}\mbox{ar}_{Y\sim \gamma_k(x,t)}\left[Y\right] = \Sigma_k(x,t). \label{eq:sigma_convergence}
\end{align}

In the additive noise setup, the distribution of $Y$ being normally distributed, the value of $U_\eta$ is actually identifiable. Let us recall the definition of the counterfactual distribution (Eq. \ref{eq:counterfactuals}) in the main text.

\begin{align*}
    P(Y_{t'}=y' \mid X=x, Y=y, T=t) = \int_u P(Y_{t'}(u)=y') P(U=u\mid X=x,T=t,Y=y)
\end{align*}

In the additive case, we have $Y\mid X=x,T=t,U_Z=u_Z = \Sigma_{u_Z}(x,t) U_{\eta} + \mu_k(x,t)$ with $U_{\eta} \sim \mathcal{N}(0,\mathbf{1})$. The counterfactual density can thus be written as

\begin{align}
    \nu_{t'}(x,t,y) &= \sum_k P(U_Z=k \mid x,y,t) \cdot \delta(u_{\eta,k}=\Sigma_k(x,t)^{-1}y - \mu_k(x,t)) \nonumber \\
    &\quad \quad \cdot \delta(y'=\Sigma_k(x,t')u_{\eta,k}+\mu_k(x,t'))
\end{align}

Considering the following estimator for the counterfactual distribution:

\begin{align*}
    \hat{\nu}_{t'}(x,t,y) = \sum_k \hat{\omega}_{k}(x,t,y) \cdot \delta(u_{\eta,k}=\hat{\Sigma}_k(x,t)^{-1}y - \hat{\mu}_k(x,t)) \cdot \delta(y'=\hat{\Sigma}_k(x,t')u_{\eta,k}+\hat{\mu}_k(x,t'))
\end{align*}

We can use Equations \ref{eq:mu_convergence}, \ref{eq:omega_convergence} and \ref{eq:sigma_convergence} to show that our estimator converges to the true distribution in the limit of infinite number of samples:

\begin{align}
    \lim_{N\rightarrow \infty} \hat{\nu}_{t'}(x,t,y) =  \nu_{t'}(x,t,y)
\end{align}

Therefore, the distance between the estimated and true counterfactual distributions reduces to 0:

\begin{align}
 \lim_{N\rightarrow \infty} W_1(\hat{\nu}_{t'}(x,t,y), \nu_{t'}(x,t,y)) = 0.
\end{align}

This concludes our proof of Result \ref{result}.

\end{document}